\documentclass[a4paper,11pt]{article}

\usepackage[utf8]{inputenc} 

\usepackage{lineno,hyperref}
\usepackage{graphicx}
\usepackage{subfigure}
\usepackage{afterpage}
\usepackage{mathrsfs} 
\usepackage{caption}
\usepackage{amsmath,amssymb,amsthm}
\usepackage{color}
\usepackage{rotating,booktabs}
\usepackage{amsfonts}
\usepackage{indentfirst}
\usepackage{mathptmx}
\usepackage{lmodern}
\usepackage{relsize,exscale}
\usepackage[doublespacing]{setspace}
\usepackage{lipsum}
\usepackage{url}
\usepackage{cancel}
\usepackage{ulem}
\usepackage{multicol}
\usepackage{calc} 
\usepackage{enumitem} 

\usepackage[T1]{fontenc}

\usepackage{amsmath,blkarray, bigstrut}
\usepackage{amsmath,array}

\usepackage{relsize,exscale}
\usepackage{algorithm2e}

\newtheorem{theorem}{Theorem}

\newtheorem{lemma}{Lemma}
\newcommand{\eqdef}{\overset{\mathrm{def}}{=\joinrel=}}

\usepackage[symbol*]{footmisc}
\DefineFNsymbolsTM{myfnsymbols}{
	\textdagger    \dagger
	\textdaggerdbl \ddagger
	\textsection   \mathsection
	\textbardbl    \|%
	\textparagraph \mathparagraph
	\textasteriskcentered *
}
\setfnsymbol{myfnsymbols}

\begin{document}

\begin{center}
	{\fontsize{14}{17} \selectfont 
		\textbf{Sorting Big Data by Revealed Preference with \\ Application to College Ranking\footnote{To cite this article: Xingwei Hu (2020): Sorting Big Data by Revealed Preference with Application to College Ranking, \textit{Journal of Big Data}, DOI: 10.1186/s40537-020-00300-1}\footnote{The views expressed herein are those of the author and should not be attributed to the IMF, its Executive Board, or its management.}}
	}

	\vspace{2cm}
	Xingwei Hu\footnote{
		Email: xhu@imf.org; \
		Phone: 202-623-8317; \
		Fax: 202-589-8317.
	}\footnote{
	The author acknowledges many helpful and detailed comments from anonymous referees.
	The author also thanks Jinpeng Ma, Kei Moriya, Bruce Moses, LLoyd S. Shapley, Wei Zhang, and seminar participants at International Monetary Fund and Stony Brook University for suggestions.
	} 
	
	\vskip .5cm
	International Monetary Fund \\
	700 19th St NW, Washington, DC 20431, USA
	\date{}
\end{center}

\newpage
\noindent \textbf{Abstract:}

\vspace{.2cm}\noindent
When ranking big data observations such as colleges in the United States, diverse consumers reveal heterogeneous preferences. 
The objective of this paper is to sort out a linear ordering for these observations and to recommend strategies to improve their relative positions in the ranking.
A properly sorted solution could help consumers make the right choices, and governments make wise policy decisions.
Previous researchers have applied exogenous weighting or multivariate regression approaches to sort big data objects, ignoring their variety and variability.
By recognizing the diversity and heterogeneity among both the observations and the consumers, we instead apply endogenous weighting to these contradictory revealed preferences.
The outcome is a consistent steady-state solution to the counterbalance equilibrium within these contradictions.
The solution takes into consideration the spillover effects of multiple-step interactions among the observations.
When information from data is efficiently revealed in preferences, the revealed preferences greatly reduce the volume of the required data in the sorting process. 
The employed approach can be applied in many other areas, such as sports team ranking, academic journal ranking, voting, and real effective exchange rates.

\vspace{.6cm}
\begin{center}
	\begin{minipage}{5.2in}
		\begin{itemize}
			\setlength\itemsep{-6pt}
			\item[\texttt{Keywords}: ]
			revealed preference;\
			authority distribution;\
			endogenous weighting; \
			college ranking;\
			big data; \
			matching game;\
			sort; \
			counterbalance equilibrium
			\item[\texttt{JEL Codes}:]
			C68,\ C71,\ C78,\ D57,\ D58,\ D74
		\end{itemize}
	\end{minipage}
\end{center}

\newpage

\section{Introduction}

\noindent
The problem we address here relates to the following typical situation: 
millions of consumers face hundreds of alternatives when making decisions. 
Each alternative is of significant complexity and variety.
From an affordable and pre-selected shortlist, a consumer picks just one alternative, thus revealing some preference over others on the shortlist.
Our objective is to sort all of these alternatives according to revealed preferences. 
One concrete context is to rank the colleges in the United States, where each student selects only one college to attend when admitted by multiple institutions.

The results in this paper arise from two considerations. 
On the one hand, college ranking is of broad public interest. 
The consumers are not only students and their families but also governments that aim to rationalize their allocation of funding as well as college administrators who are interested
in comparing their ranking to peer institutions  on a national level. 
Alumni pay attention to similar comparisons, and  leading companies prefer to hire graduates from the best colleges.
As a consequence, there are dozens of college rankings in the U.S. market, including those published by mainstream media, such as U.S. News \& World Report, 
Washington Monthly, the Wall Street Journal (WSJ), Time, and Forbes.
As indicated by many researchers (e.g., Bastedo and Bowman, 2010 and 2011), these rankings have demonstrable effects on potential students and administrators due to their perceived influence on resource providers.
On the other hand, however, almost all these rankings rely on some preset weighting system using merely several criteria, such as acceptance rates, standardized test scores, alumni donation rates, and class sizes.
The weights are then applied to each college to obtain a ranking score. 
The selection of criteria and the weights are subjective; a slight variation leads to a different ranking result.
A consequence of this methodology is an isomorphism: the diversity of the higher educational institutions are not valued. 
Many researchers have claimed such rankings often harm higher education (e.g., Craig, 2015; Moed, 2017; Perez-Pena and Slotnik, 2012).

Our solution to the problem is to apply authority distribution (Hu and Shapley, 2003) to the revealed preferences by the students.
We believe there  exists no exogenous weighting system applicable to all colleges. 
Each college is unique, and just a few measurements cannot accurately portray it.
For example, a liberal arts college and an engineering-focused college could value SAT scores and faculty publications quite differently.
The characteristics of a college are embodied in the applicants in general, and the admitted students in particular.
We remove the subjective weighting systems and ignore the criteria selection. 
However, we believe each consumer could have his or her specific weighting on a much broader set of criteria, including qualitative and latent ones.
 A student could have dozens of other considerations, such as the distance from home, tuition, sports, job and internship potential, personal connections, campus visit experience, and median salary after graduation ---far beyond the few criteria listed in mainstream media.
Personal considerations and weighting are effectively revealed in his or her preference when selecting one college and rejecting others.
Identifying logical and consistent inferences from these personal preferences is the main challenge.
 Hu and Shapley (2003) partially overcomes the challenge and also includes a simple college ranking using artificial data. 
In this paper, we build a detailed roadmap to get around the obstacles in ranking colleges.

 The approach has three advantages.
First, we allow heterogeneity from the consumers as well as from the colleges. 
We use a massive endogenous weighting scheme to avoid any subjective weighting and criteria selection.
Secondly, our method is robust. 
We use revealed preferences  as a dimensionality reduction mechanism ---each preference could have taken into consideration countless selection criteria, and the ranking uses preferences from millions of households.
Additionally, our ranking considers long-term and system-wide influences by aggregating spillover effects from direct bilateral influences.
Thirdly, the specific application to the game-theoretic authority distribution respects individual rationality and values collective welfare. 
It seamlessly links authority distribution with revealed preference, heterogeneity, big data, endogenous weighting, strategyproofness, and public interests in college ranking. 
Furthermore, we envision many potential applications in other fields.
As a consequence of these advantages, we believe our ranking methodology is more authoritative than those used in the popular rankings.

We organize the remainder of the paper as follows. 
Section \ref{sect:methodology} summarizes the critical issues in other rankings and introduces the new sorting methodology, which integrates the revealed preference  with the authority distribution framework.
Section \ref{sect:EST} analyzes the data, calculates the ranking scores, and simulates their confidence intervals.
 Next, Section \ref{sect:Property} discusses a few properties of the ranking method.
Notably, we address strategies to improve ranking scores.
Section \ref{sect:empirical} studies an empirical ranking using real data.
 In Sections \ref{sect:discussion}, we discuss vulnerabilities, extensions, and other applications of the framework.
Finally, we conclude in Section \ref{sect:conclusion}.
Our exposition is self-contained, and the proofs are in the Appendix.

\section{The Methodology}\label{sect:methodology}
\noindent
Before our formal discussion, we introduce a few notations.
Let $\mathbb{N} = \{1, 2,\cdots,n\}$ denote the set of colleges in the United States, indexed as $1,2,..., n$. 
In the time frame of one academic year, let $A_i$ be the number of students admitted by college $i$. 
Out of the $A_i$ admitted students, $\tilde N_{ij}$ students would attend college $j \in \mathbb{N}$.
We use a multinomial distribution to model these random numbers $\tilde N_{ij}$, using the parameters $A_i$ and $(P_{i1},\cdots, P_{in})$.
The probability $P_{ij}$ is the likelihood that anyone in $A_i$ would enroll in college $j$.
Of course, the student is also admitted by college $j$.
We write matrix $ P = \left [ P_{ij} \right ], i,j=1,2,\cdots, n$.
After the admission and acceptance process has been completed, we observe a $\tilde N_{ii}$, which is the number of enrolled students, denoted by $E_i$.
For $j\not = i$, however, $\tilde N_{ij}$ are generally not observable at this time  or thereafter.
The data $A_i$ and $E_i$ are available  from many sources, including the National Center for Education Statistics (NCES) and the Common Data Set (CDS) for these colleges.

\subsection{Key Issues in Popular Rankings}
\noindent
 Of the dozens of popular college rankings in the United States, each uses several criteria to weight the colleges and calculate a weighted score for each college. 
For example, the following seven criteria drove the U.S. News \& World Report's college ranking ( U.S. News) in 2019:
Graduation and Retention Rates (22.5\%),
Undergraduate Academic Reputation (22.5\%),
Faculty Resources (20\%),
Student Selectivity (12.5\%),
Financial Resources (10\%),
Graduation Rate Performance (7.5\%),
and Alumni Giving Rate (5\%).
This weighting system raises several questions: Why are other criteria not included? How were the weights determined? How do we measure qualitative variables such as selectivity and reputation? 
 Also, ``giving rates'' and ``resources'' data are vulnerable to manipulation.

A generic recipe for these rankings is as follows.
Let there be $k$ criteria, $w_j$ be the weight on the $j$th criterion, and $x_{ji}$ be college $i$'s score on the $j$th criterion.
These rankings use 
\begin{equation}\label{eq:popular_weighting}
\left (
w_1, \cdots, w_k
\right )
\left [
\begin{array}{ccc}
x_{11} & \ldots & x_{1n} \\
\vdots & \vdots & \vdots \\
x_{k1} & \ldots & x_{kn}
\end{array}
\right ]
=
\left (
\sum\limits_{j=1}^k w_j x_{j1},
\cdots,
\sum\limits_{j=1}^k w_j x_{jn}
\right )
\end{equation}
to calculate a ranking score $\sum\limits_{j=1}^k w_j x_{ji}$ for college $i$. The ranking scores on the right side of (\ref{eq:popular_weighting}) are then used to sort the colleges. 
Variants of criteria and their weights can produce quite different ranks for the same college. 
Table \ref{tab:UCLA_Ranks} is an example: the lowest rank of $31$ for UCLA (University of California, Los Angeles) is six times more than the highest rank of $5$ for the same college.

\begin{table}
\caption{Ranks for the Same College: the UCLA Case$^*$}
\label{tab:UCLA_Ranks}
\centering
\footnotesize
\begin{tabular}{r || c | c | c | c | c | c} \hline \hline
   &  &   &U.S. News&Business &WSJ / &Washington\\
Source&ARWU&Forbes&Best Colleges&Insider &Times HE &Monthly  \\ \hline
Rank &9  &31  &19   &22    &12  &8     \\ \hline \hline
	 &  &   &U.S. News&Princeton&Avery et al.&     \\ 
Source&QS &Niche &Global Universities &Review  &(2013)&Parchment \\ \hline 
Rank &16 &26  &9   &5    &28  &18    \\ \hline \hline
\multicolumn{7}{l} {* : ARWU and QS for Academic Ranking of World Universities and Quacquarelli Symonds, resp.}\\
\multicolumn{7}{l} {* : From websites cited in the reference, accessed 18 March 2019.} \\
\multicolumn{7}{l} {* : Liberal arts colleges and military academies are excluded.}\\
\end{tabular}
\end{table}

However, these popular rankings have been widely criticized (e.g., Bruni, 2016; Ehrenberg, 2005; Grewal, Dearden, and Lilien, 2008; Luca and Smith, 2011; Moed, 2017; Perez-Pena and Slotnik, 2012). 
Let us mention just a few reasons why.
First, the premise of these popular rankings ---that a few criteria will produce a complete comparison for all colleges--- is not universally shared. 
It does not offer a realistic scenario nor a good approximation of one. 
Each college has not only a unique location but also unique features.
Some , for example, focus on research while others on teaching. 
Some are science- and engineering-based, while others are arts- and social science-oriented.
Secondly, these rankings hardly align with the selection criteria of consumers.
In particular, they ignore several vital concerns of prospective students,
such as the distance between college and home, campus life, sports, median post-graduation starting salaries, location, and the number of graduate programs. 
 Of course, tuition and room and board are among the most significant concerns for students who apply for education loans.
All of these considerations are taken into account when choosing one college out of many.
Thirdly, these rankings inherit the subjectivity and bias from a single model designed by a small committee of educators and business people.
For example, the weights $w_j$ in (\ref{eq:popular_weighting}) are very subjective; many scores $x_{ji}$ ---such as academic prestige--- are also subjective,
and they are subject to manipulation or inflation by colleges.
Lastly, these criteria-based rankings discourage institutional diversity and reinforce similarities among universities.
If colleges were to base their development plans on the U.S. News criteria in order to raise their ranks, there might be little value in retaining programs such as sports.

A recent ranking by Avery et al. (2013) is worth highlighting. 
In their research, they study several econometric models that describe the decision process by colleges versus applicants. 
An extensive set of data was collected to estimate the unknown weights in these models. 
The result is three sets of rankings.
In contrast, we study college versus college comparisons, using rejection and acceptance by the students as the basis for comparison. 
We believe that sorting or ranking is about comparisons, and anyone's preference is already a ranking between two colleges by that individual.
Also, we argue that a constant coefficient in an econometric model cannot capture the marginal effect for all schools or all students. 
If a regression model is used, then its residual term accounts for the distinctive characteristics of a college or the unique considerations of a student.
Unfortunately, the residual term is generally treated as random noise.
Our new ranking method, in contrast, ingests every piece of recent public information about a college, including any scandals and victories by the football team, for example.

In contrast to (\ref{eq:popular_weighting}), we write a deterministic decision function $F_{is}(\cdots)$ for student $s$' choice $y_{is}$ on college $i$,
\begin{equation}\label{eq:BigData}
y_{is} = F_{is}(x_1, x_2, x_3, \cdots)
\end{equation}
where $(x_1, x_2, x_3, \cdots)$ is an indefinite list of all possible considerations for all of the students.
The big data $(x_1, x_2, x_3, \cdots)$ could include thousands of explanatory variables about the colleges.
Being viewed as an observation of available data  on the Internet, a college could be as ``big'' as Harvard.
However, these functions $F_{is}(\cdots)$ are personal-specific and not identifiable, given the availability of the data. 
Similarly, we could also define an indefinite-dimensional decision function for each college.
  Because of the dimensionality of $(x_1, x_2, x_3, \cdots)$ and the privacy of $F_{is}(\cdots)$, some level of data abstraction becomes indispensable to order the colleges from the best to the worst.
In the literature, however, there is hardly any research which targets sorting big data objects, though
many (e.g., Eeckhout, Pinheiro, and Schmidheiny, 2010 and 2014; Dohmen and Falk, 2011; Langville and Meyer, 2012) have sorted complicated objects using spatial, network, or multidimensional analyses.

\subsection{Revealed Preference}
\noindent
Revealed preference theory (cf. Samuelson, 1948) analyzes the actual choices by individual consumers who have heterogeneous utility functions. 
An individual may execute individual rationality or bounded individual rationality by maximizing his or her utility function.
Individual consumers may also demonstrate behavioral biases in making their choices.
Revealed preference, however, does not hold the axioms of a utility function or a multi-person utility function (Baucells and Shapley, 2008).
The focus is on what choice $y_{is}$ individual $s$ makes, not on how he or she makes the choice.
Thus, we ignore the personal decision function $F_{is}(\cdots)$ and the explanatory variables $(x_1, x_2, x_3, \cdots)$ in (\ref{eq:BigData}).
The choice $y_{is}$ has considered them.
Personal considerations, however, play an essential role in making an individual choice. 
A typical practice for a student is to compare one college with another without knowing or using any utility function.
Thus, a universally accepted set of considerations and decision functions does not work for millions of high school seniors. 
 In the literature, revealed preference analysis had been used as an alternative to regression approaches (e.g., Chirinko and Schaller, 2004; Pritchard, 2018; Tieskens et al., 2018).

 By ignoring $F_{is}(\cdots)$ and $(x_1, x_2, \cdots)$, we transform the big data into a high-dimensional space of $y_{is}$.
In the transformation, the underlying assumption is that personal preferences have adequately reflected the ignored data.
A significant reason for the assumption is the informational efficiency in the preference revelation.
The efficiency is driven by the strong motivation of parents and students to acquire relevant information.
 To gather that data, they might fly thousands of miles for a campus visit or spend weeks in examining slight differences between two candidate colleges.
The efficiency is also due to the collective power to assemble and assimilate the information by millions of consumers, 
though an  individual is influenced by emotional biases and could make cognitive errors in processing the information.

 Colleges vary significantly among the higher education system.
A college is also a rational decision-maker that admits the best students it can attract. 
Its preference is effectively revealed in its admitted students.
Accordingly, there exists no universally accepted set of criteria for all these colleges. 
For example, a research-focused college may have a different utility function, if there is any, from a teaching-oriented college.
A public university has to comply with specific enrollment mandates set by state authorities.
The mandates and public spending on higher education are also not acknowledged in the rankings by mainstream media.
Admission procedures vary among colleges.
Unlike the idealized selection process stated in Gale and Shapley (1962), a college cannot rank all of its applicants, and it generally offers more admissions than needed.
It may establish a waiting list, and it may also offer early admissions and scholarships in order to hold a repository of students who can no longer accept offers from other schools.

In the many-to-many matching game between the students and the colleges, preference is officially revealed at least three times.
First, a student files applications to a shortlist of pre-selected schools, ignoring any unfavorite or unfit ones.
A student could take years to observe and consider the shortlist before filing applications.
Preference is revealed a second time when a college admits potential students who satisfy the standards of the college.
 A college admission officer also considers enrollment size and diversity of the admitted students. 
In the last stage, the student decides on his or her favorite school and rejects all others.

Lastly, these revealed preferences are contradictory; they offer no utility function nor any linear ordering of the alternatives.
Indeed,  neither students  nor parents use a utility function or an econometric model to make decisions.
They make pairwise comparisons.
 Nevertheless, we cannot apply any comparison sorting algorithms for arrays, such as Quicksort or Heapsort; revealed preferences also show how much one alternative is preferred over another (cf. Table \ref{tab:SamplePreference}).
This paper applies authority distribution
to combine these conflicting preferences, which gradually spills over the direct influence among all colleges.
The evolution of the spillover effects eventually smooths out all conflicts among the preferences. 
The infinite-step evolution process can be simplified by solving a counterbalance equilibrium equation in (\ref{eq:authority_distribution}).

\subsection{Authority Distribution}
\noindent
In the context of authority distribution, there is a network of multiple players. 
These players have a direct influence, broadly called \textit{power}, over each other. 
For each player, the network constitutes a ``command game'' by the direct influence on the player. 
 Together, there are $n$ command games that describe the power-in-and-out dynamics in the network.
When the players are sufficiently connected, there exists a general equilibrium, called counterbalance equilibrium, which is the steady-state solution for the power flow dynamics.
A player derives authority from others, whom he directly influences. 
When we apply authority distribution to sort the U.S. higher education institutions, the players are the colleges.
The direct influence could be, for example, the acceptance and rejection by prospective students, sports,
faculty recruitment, research fund competition, and publication cross-citation. 
In this paper, we focus solely on the preferences of potential students who make rational choices after observing the other aspects of influence;
we argue that consumers' choices make the most comprehensive interactions when comparing two choices.
In the literature, researchers (e.g., Bastedo and Bowman, 2011; Correntea, Grecoa, and Lowinskic, 2018; Grewal, Dearden, and Lilien, 2008) have also recognized the interdependence of institutions when evaluating colleges.

For college $i$, for example, its pool of admitted students and their choices constitute a game of foot voting in which students' preferences are explicitly exercised through their actions of acceptance or rejection, i.e., $y_{is}$. 
In this student-college matching, dollar voting is part of the foot voting, as tuition is a significant concern for parents and student borrowers.
Ironically, by excluding tuition in their criteria, mainstream media ignore one of the most salient issues among  education consumers.
Besides, the pool itself represents college $i$'s characteristics: it could be competitive or not; it could be small or large based on its enrollment size.
Statistics about GPAs, standardized test scores, and demographic data are already embedded in the pool of admitted students.

\begin{figure}
\centering
\begin{tabular}[b]{c}
\includegraphics[height=3.5cm, width=4.5cm]{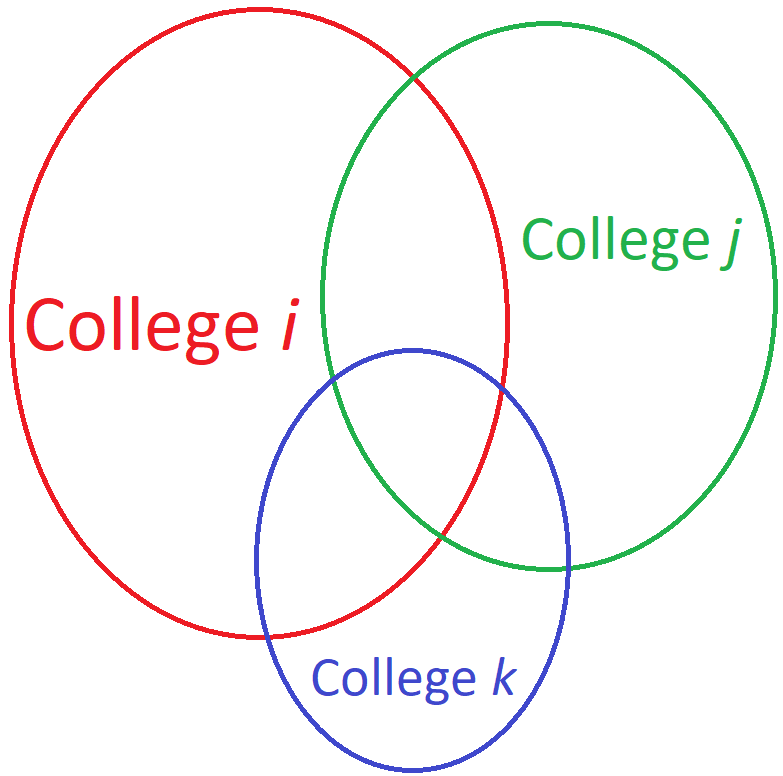} \\
\small (a) Shared Pools of Admitted Students
\end{tabular} \hspace{.2cm}
\begin{tabular}[b]{c}
\includegraphics[height=3.5cm, width=4.5cm]{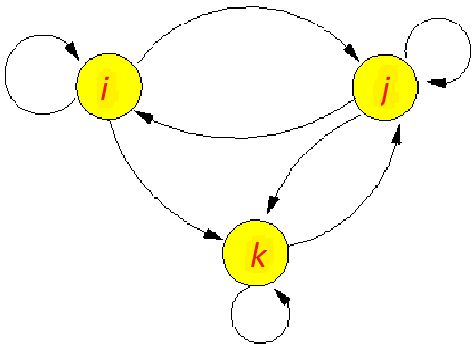} \\
\small (b) Power Transition Dynamics
\end{tabular}
\caption{Interactions Among Colleges $i$, $j$, and $k$.}
\label{fig:3Colleges}
\end{figure}

Colleges $i$ and $j$ become rivals to compete in recruiting the shared pool of students when their pools overlap.
Each college has its rivals, and the number of rivals varies across the higher education system. 
Additionally, two non-competing colleges may have one or more common rivals, and
three or more pools could also intersect (see Figure \ref{fig:3Colleges}(a)).
Consequently, a competitive battlefield of such a shared pool could involve many players.

For all the pools, the power transition matrix $P$ collectively specifies the likelihood of any student's final choice after admission.
As the proportion of students who decide to attend college $j$, 
$P_{ij}$ measures college $j$'s direct influence or power in college $i$'s foot voting game.
Asymmetry generally exists when $P_{ij}\not = P_{ji}$; thus, power flows from one college to another unevenly.
 Figure \ref{fig:3Colleges}(b) shows the bilateral power movement among three colleges; each has a direct influence on the other two. 
 Besides, when aggregating $y_{is}$ to the matrix $P$, we further reduce the dimension of data.

Authority distribution associated with $P$ is a row vector $\pi = (\pi_1, \pi_2, \cdots, \pi_n)$ which satisfies the counterbalance equilibrium
\begin{equation} \label{eq:authority_distribution}
(\pi_1, \pi_2, \cdots, \pi_n) = (\pi_1, \pi_2, \cdots, \pi_n) P
\end{equation}
subject to the normalization condition $\sum\limits_{i=1}^n\pi_i = 1$ and positivity condition $\pi_i \ge 0$ for all $i\in \mathbb{N}$.
We can see a direct influence from two aspects.
For a specific $i\in \mathbb{N}$, it derives authority from others on whom it has a direct impact, i.e., by (\ref{eq:authority_distribution}),
\begin{equation} \label{eq:inflow}
\pi_i = \sum\limits_{j=1}^n \pi_j P_{ji}.
\end{equation}
Equation (\ref{eq:inflow}) describes how power flows from other players into $i$.
It derives more authority from influential players than from non-influential players, other things being equal. 
Moreover, it also derives more authority from players on which it has a large direct influence, other things being equal. 
In the second aspect, $i$ also contributes to other colleges that have direct influences on $i$. This can be seen from the power outflow equation
\begin{equation} \label{eq:outflow}
\pi_k =  \pi_i P_{ik} \ + \ \sum\limits_{j\not = i}\pi_j P_{jk}. 
\end{equation}
The larger $P_{ik}$, the more college $i$ contributes to college $k$'s authority.
The inflow power in (\ref{eq:inflow}) and outflow power in (\ref{eq:outflow}) eventually reach an equilibrium described by (\ref{eq:authority_distribution}).
In essence, $\pi$ is a power index in the interactive and yet controversial network.
The derivation of the counterbalance equilibrium and the properties of $\pi$ were studied in Hu and Shapley (2003). 
This distribution has been mostly used in social networks, corporate networks, and controls (e.g., Crama and Leruth, 2007; Grabisch and Rusinowska, 2010).
 Compared to Google's PageRank (e.g., Langville and Meyer, 2012), authority distribution drops the damping factor and has a stochastic matrix with a non-zero diagonal.

\section{Estimation} \label{sect:EST}
\noindent
This section estimates $P$, $\pi$, and its confidence interval, mean, and median.

\subsection{The Data}
\noindent
All the data used to estimate $P$ come from the Internet.
With regard to sampling errors, official admission data provide high precision to the diagonal elements of $P$ while online survey data cast more doubt on the off-diagonal ones.

\subsubsection{The Official Enrollment Rates}
\noindent
The probability $P_{ii}$ is the likelihood that any student admitted by college $i$ would enroll in college $i$.
We can estimate it by the enrollment rate
\begin{equation}\label{eq:enrollment_rate}
\hat{P}_{ii} = \frac{E_i}{A_i}.
\end{equation}
The enrollment rates are available on many websites, including the CDS, the NCES, Wikipedia, U.S. News, and those for colleges.
Table \ref{tab:UCLA_Admission_Stat} is an example of admission statistics for UCLA between 2013 and 2018. 
In this table, UCLA admitted 15,988 of 113,779 applicants for fall 2018. Out of the admitted students, 6,240 enrolled at UCLA. 
Thus, the enrollment rate was $\frac{6,240}{15,988} = 39.03\%$ for 2018.

\begin{table}
\caption{UCLA Fall Admission Statistics 2013-2018$^*$}
\label{tab:UCLA_Admission_Stat}
\centering
\small
\begin{tabular}{r || c | c | c | c | c | c} \hline \hline
				&2018  &2017  &2016 &2015 &2014 &2013 \\ \hline
Applicants   &113,779&102,242&97,115&92,728&86,548&80,522\\
Admits ($A_i$) &15,988 &16,456 &17,474&16,016&16,059&16,448\\
Enrolled ($E_i$)&6,240 &6,038 &6,545 &5,679 &5,764 &5,697 \\ \hline \hline
\multicolumn{7}{l}{* Source : \url{www.wikipedia.org},  accessed 1 January 2019.}
\end{tabular}
\end{table}

\subsubsection{The Survey Data of Preference}
\noindent
When $i\not=j$, $P_{ij}$ is the percentage of students who are admitted by college $i$ but  decide to attend college $j$. 
 These data  are not wholly available, so we estimate them from the survey data of preference posted on Parchment, Niche, and similar websites.
Many students also post their decisions on online discussion boards.
Table \ref{tab:SamplePreference} lists a sample of revealed preferences, described as odds ratios, for twelve colleges. These include seven private universities and five public ones, often ranked highly by mainstream media. 
At the $i$th row and $j$th column, the numerator is the percentage for college $i$ and the denominator for college $j$.
For example, of the students admitted by both Harvard and Stanford and deciding to attend either Harvard or Stanford, 44\% choose Stanford, and 56\% Harvard.
Also, the percentage 56\% has the .95 confidence interval between 51.2\% to 60.7\%, listed on the Parchment website, and calculated by the Wilson score method (Wilson, 1927). 
Similarly, the percentage 44\% has the .95 confidence interval between 39.3\% to 48.8\%.
Of course, these students may also be admitted by other schools.

\begin{table}
\caption{Sample Revealed Preference Represented as Odds Ratios$^*$}
\label{tab:SamplePreference}
\centering
\small
\fontsize{8}{9}\selectfont
\begin{tabular}{r | c | c | c | c | c | c | c | c| c| c | c } \hline \hline
	&HRD      &SFD      &Yale      &PRT      &MIT      &CCG      &CIT      &UCB      &Mich      &UCLA      &UVA      \\ \hline
SFD &$\frac{44}{56}$&        &        &        &        &        &        &        &        &        &        \\
Yale&$\frac{37}{63}$&$\frac{45}{55}$&        &        &        &        &        &        &        &        &        \\
PRT &$\frac{24}{76}$&$\frac{30}{70}$&$\frac{33}{67}$&        &        &        &        &        &        &        &        \\
MIT &$\frac{38}{62}$&$\frac{32}{68}$&$\frac{64}{36}$&$\frac{55}{45}$&        &        &        &        &        &        &        \\
CCG &$\frac{30}{70}$&$\frac{32}{68}$&$\frac{24}{76}$&$\frac{38}{62}$&$\frac{26}{74}$&        &        &        &        &        &        \\
CIT &$\frac{16}{84}$&$\frac{25}{75}$&$\frac{45}{55}$&$\frac{44}{56}$&$\frac{19}{81}$&$\frac{46}{54}$&        &        &        &        &        \\
UCB &$\frac{20}{80}$&$\frac{10}{90}$&$\frac{10}{90}$&$\frac{16}{84}$&$\frac{14}{86}$&$\frac{20}{80}$&$\frac{24}{76}$&        &        &        &        \\
Mich&$\frac{16}{84}$&$\frac{15}{85}$&$\frac{22}{78}$&$\frac{18}{82}$&$\frac{9}{91}$ &$\frac{29}{71}$&$\frac{20}{80}$&$\frac{39}{61}$&        &        &        \\
UCLA&$\frac{18}{82}$&$\frac{6}{94}$ &$\frac{15}{85}$&$\frac{17}{83}$&$\frac{9}{91}$ &$\frac{21}{79}$&$\frac{8}{92}$ &$\frac{47}{53}$&$\frac{43}{57}$&        &        \\
UVA &$\frac{25}{75}$&$\frac{13}{87}$&$\frac{21}{79}$&$\frac{6}{94}$ &$\frac{17}{83}$&$\frac{13}{87}$&$\frac{13}{87}$&$\frac{36}{64}$&$\frac{56}{44}$&$\frac{48}{52}$&        \\
UNC &$\frac{20}{80}$&$\frac{28}{72}$&$\frac{21}{79}$&$\frac{11}{89}$&$\frac{33}{67}$&$\frac{33}{67}$&$\frac{40}{60}$&$\frac{36}{64}$&$\frac{45}{55}$&$\frac{45}{55}$&$\frac{47}{53}$\\ \hline \hline
\multicolumn{12}{l}{* HRD, SFD, PRT, CCG, CIT are for Harvard, Stanford, Princeton, Chicago, CalTech, resp.} \\
\multicolumn{12}{l}{* UCB and Mich are for University of California at Berkeley and University of Michigan, resp.} \\
\multicolumn{12}{l}{* UVA and UNC are for University of Virginia and University of North Carolina, resp.} \\
\multicolumn{12}{l}{* Source : \url{www.parchment.com}, accessed 1 November 2019.} \\
\end{tabular}
\end{table}

 Some simple conclusions could be reached by using only the data in Table \ref{tab:SamplePreference}. 
First, column 1 shows that Harvard is preferred over all other colleges; a reasonable ranking would place it before the others, regardless of the confidence intervals of or the weights on the odds ratios.
Secondly, we would expect Stanford to place second since it is preferred over all others, except Harvard (cf. Column 2). 
It is also at the top when Harvard acts as the reference (cf. Column 1).
A further analysis implies that MIT and Yale would compete for third place.
They are preferred over all other schools except Harvard and Stanford (cf. Columns 3 and 5).
When either Harvard or Standford acts as the reference (cf. Columns 1 and 2), they place higher than all other schools except Stanford and Harvard, respectively.

However, there are a few data issues worth mentioning, as different odds ratios come from different shared pools of students.
First, the transitivity of preference does not hold. 
For example, we could find a circular chain of preference, Mich $\preccurlyeq$ UVA $\preccurlyeq$ UCLA $\preccurlyeq$ Mich, when using the bilateral preferences only. 
Secondly, the ratios are not multiplicative. 
For instance, UCLA $\preccurlyeq$ Mich by $\frac{43}{57}$ and UCLA $\preccurlyeq$ UCB by $\frac{47}{53}$ does not imply UCB $\preccurlyeq$ Mich by $\frac{53}{47} \frac{43}{57} = \frac{2,279}{2,679}$. 
On the contrary, Mich $\preccurlyeq$ UCB by $\frac{39}{61}$.
Thirdly, contradictory orderings exist if we compare colleges using different benchmarks. 
For example, if we use Harvard as the reference benchmark (cf. Column 1), then Mich $\preccurlyeq$ UCLA $\preccurlyeq$ UVA. 
But if we use UCB as the reference school (cf. Column 8), the ordering is UVA $\preccurlyeq$ Mich $\preccurlyeq$ UCLA.
If Harvard has more authority than UCB, then the first ordering should be weighted more than the second one.
Also, spatial adjacency plays a crucial role in these odds ratios. 
In the seventh row, for example, Berkeley has the lowest odds ratio versus Stanford, and its distance from Stanford is also the shortest.
A similar case is the seventh column, where UCLA has the lowest odds ratio against CalTech, and the shortest distance from CalTech.
Lastly, the odds ratios alone are not enough to determine all $P_{ij}$; we are more interested in the number of students who are admitted by college $i$ but decide to go to college $j$.

\subsection{Conditional Estimability of $P$} \label{subsect:Estimate_P}
\noindent
 This subsection illustrates how to estimate $P_{ij}$, conditional on $P_{ii} = \frac{E_i}{A_i}$ and other observations.
Clearly, the enrollment rate $\frac{E_i}{A_i}$ is an unbiased estimate for $P_{ii}$. 
Besides, both $E_i$ and $A_i$ from the CDS and the NCES have an annual frequency.
But the preference survey data have a real-time frequency, and they likely come from multiple years. 
In order to fix the frequency divergence, we assume that both the odds ratios and the enrollment rates remain stable across multiple years.

 To link the odds ratios in Table \ref{tab:SamplePreference} to $P$, we capitalize on the confidence intervals of the odds ratios.
Out of the students who are admitted by both colleges $i$ and $j$ and decide to attend either $i$ or $j$, we let $S_{ij}$ be the likelihood a student would choose college $j$. 
We also let $M_{ij}$ be the number of those students who participate in the online preference survey and
let $s_{ij}$ be the proportion of the surveyed students who decide on college $j$.
 Clearly, $M_{ij}=M_{ji}$ and the odds ratio is $\frac{1-s_{ij}}{s_{ij}}$.
We define $\omega_{ij} = \frac{2 s_{ij} M_{ij} + z^2}{2 M_{ij} + 2 z^2}$ where $z$ is the $.975$ percentile of the standard normal distribution.
Given $s_{ij}$ and $M_{ij}$, the $.95$ confidence interval for $S_{ij}$, calculated by the Wilson score method, is
\begin{equation}\label{eq:wilson}
\omega_{ij}\ \pm \ \frac{z}{2 M_{ij} + 2 z^2} \sqrt{ 4 s_{ij}(1-s_{ij}) M_{ij} + z^2}.
\end{equation}
Replacing $s_{ij}$ with $\frac{E_i}{A_i}$ and $M_{ij}$ with $A_i$ in (\ref{eq:wilson}), we obtain the Wilson confidence interval for $P_{ii}$.
Given $s_{ij}$ and the confidence interval (\ref{eq:wilson}), we can extract $M_{ij}$ by Lemma \ref{lm:tilde_M_ij}.
Alternatively, we can also extract $M_{ij}$ using $s_{ij}$ and the confidence interval length 
$\frac{z}{M_{ij} + z^2} \sqrt{ 4 s_{ij}(1-s_{ij}) M_{ij} + z^2}$.

\begin{lemma}\label{lm:tilde_M_ij} 
Let $\tau = 1 - s_{ij}$ and let $\eta$ be the lower confidence bound in (\ref{eq:wilson}). Then
$$
M_{ij} 
= 
\frac{\tau s_{ij}-(s_{ij}-\eta)(1-2\eta) + \sqrt{[(s_{ij}-\eta)(1-2\eta)-\tau s_{ij}]^2 + 4\eta (1-\eta) (s_{ij}-\eta)^2}}
{2\left (\frac{s_{ij}-\eta}{z} \right )^2}.
$$
\end{lemma}

Thus, $N_{ij} \eqdef s_{ij} M_{ij}$ is the number of students in the survey who are admitted by college $i$ but decide to attend college $j$.
Consequently, $\sum\limits_{k \not = i} N_{ik}$ is the number of students in the survey who are admitted by college $i$ but decide not to attend college $i$.
This is about $1-P_{ii}$ of the students in the survey who are admitted by college $i$, if  each student has the same likelihood to take the survey.
Therefore, we estimate $P_{ij}$ by
\begin{equation} \label{eq:Pij_Estimation}
\hat P_{ij} \ \eqdef \ \frac{N_{ij}}{\sum\limits_{k\not = i} N_{ik}} \left (1 - \frac{E_i}{A_i} \right ), \qquad j\not = i.
\end{equation}
As stated in Theorem \ref{thm:expected_pij}, $\hat P_{ij}$ is an unbiased estimate for $P_{ij}$ given $P_{ii} = \frac{E_i}{A_i}$, the size of the survey data $\sum\limits_{k \not = i} N_{ik} $, and certain reasonable assumptions.
\begin{theorem}[Conditional Unbiasedness]\label{thm:expected_pij} 
Assume all students accepted by college $i$ have the same likelihood to independently participate in the preference survey. 
Then
$$
\mathrm{E} \left [ \hat P_{ij} \left | \right . P_{ii} = \hat P_{ii}, \sum\limits_{k \not = i} N_{ik} \right ] = P_{ij}, \quad \forall \ j\not=i.
$$
\end{theorem}

 As the breadth and size of each college differ significantly across the education system, we mitigate the size effect in the enrollment without modifying the odds ratios.
The above estimation of $P_{ij}$ for all $j\not = i$ is based on unequal sample sizes.
The size of $M_{ij}$ relies on the size of $A_j$, but $A_j$ varies across all $j\not = i$.
To make $P_{ij}$ and $P_{ik}$ comparable roughly under a common sample size, we scale the length of the Wilson confidence interval (\ref{eq:wilson})
by $\sqrt{\frac{E_j}{\max\limits_{k\not =i} E_k}}$, i.e., the scaled Wilson confidence interval is
\begin{equation}\label{eq:new_wilson}
\omega_{ij}\ \pm \ \frac{z}{2 M_{ij} + 2 z^2} \sqrt{ 4 s_{ij}(1-s_{ij}) M_{ij} + z^2} \ \sqrt{\frac{E_j}{\max\limits_{k\not =i} E_k}}.
\end{equation}
Before the scaling, the length is of order $O\left ( \frac{1}{\sqrt{M_{ij}}} \right )$. 
After the scaling, it is of $O \left ( \sqrt{ \frac{E_j}{M_{ij} } }\right )$, 
ignoring the common denominator $\sqrt{\max\limits_{k\not =i} E_k}$.
If {$\frac{E_j}{M_{ij}}$} is close to a non-zero constant as $E_j\to \infty$, then the length of the scaled interval is 
of $O\left ( \frac{1}{\sqrt{\max\limits_{k\not =i} E_k}}\right )$ and 
the common sample size is of $O \left ( \max\limits_{k\not =i}  E_k \right )$.
As $P_{ij}$ only counts the students enrolling in college $j$, we  use $E_j$ in (\ref{eq:new_wilson}) to exclude the students who do not enroll in college $j$.
 Using $\max\limits_{k\not =i} E_k$ guarantees that the scaled intervals (\ref{eq:new_wilson}) lie in $(0,1)$.
The scaling by a constant is different from the actual increase of the sample size; the latter extracts more information from new data and thus reduces the uncertainty in estimation.
Scaling by the square root of sample size is a common practice to balance the breadth and depth, e.g., $t$-statistic with unequal sample size, the investor's breadth, and Grinold and Kahn (2011).
In summary, Algorithm \ref{alg:P_ij} estimates $P_{ij}$ when the enrollment sizes $E_j$ have a large variation:
\begin{figure}[htb]
\centering
\begin{minipage}{.85\linewidth}
\begin{algorithm}[H]\label{alg:P_ij}
\SetAlgoLined
1. Calculate the Wilson interval (\ref{eq:wilson}) if it is not available\;
2. Scale the Wilson confidence interval using (\ref{eq:new_wilson})\;
3. Apply Lemma \ref{lm:tilde_M_ij} to (\ref{eq:new_wilson}) to calculate the scaled $M_{ij}$ and $N_{ij}$\;
4. Use the scaled $N_{ij}$ in (\ref{eq:Pij_Estimation}) to calculate $\hat P_{ij}$.\
\vspace{.5cm}
\caption{Estimate $P_{ij}$ for $j\not= i$.}
\end{algorithm}
\end{minipage}
\end{figure}

\subsection{Confidence Interval of $\pi$}
\noindent
After estimating $P$, we solve the counterbalance equation $\hat \pi = \hat \pi \hat P$ by Algorithm \ref{alg:Solve_pi} , where $\mathbf{1}_n$ is the column vector with $n$ ones.
According to the theory of Markov chains, $\pi^{(t)}$ converges if $\hat P$ satisfies certain properties specified in Theorem \ref{thm:invaniance_direct_indirect}. 
After setting the final $\pi^{(t)}$ to $\hat \pi$,
we sort the vector $\hat \pi$ from the largest value to the least. 
The college with the largest value ranks first, and the college with the least value ranks last. 
\begin{figure}[htb]
\centering
\begin{minipage}{.55\linewidth}
\begin{algorithm}[H]\label{alg:Solve_pi}
\SetAlgoLined
$\pi^{(0)} \longleftarrow \mathbf{1}_n'$; \ $\pi^{(1)} \longleftarrow \frac{1}{n} \mathbf{1}_n'$; \ $t \longleftarrow 1$\; 
\While{ $|| \pi^{(t)} - \pi^{(t-1)}||_\infty> 1e-9$ }{
	$t \longleftarrow t+1$\;
	$\pi^{(t)} \longleftarrow \pi^{(t-1)} \hat P$\;
}
\vspace{.5cm}
\caption{Calculate $\hat \pi$ from $\hat P$.}
\end{algorithm}
\end{minipage}
\end{figure}

The uncertainty of $\hat \pi$ comes from the estimation of $P$. 
We can capitalize on the Monte Carlo simulation to estimate the confidence interval of $\hat \pi$,
by simulating $20,000$ power transition matrices $P$.
For each simulated $P$, we calculate a ranking score $\hat \pi$.
From these $20,000$ sets of ranking scores $\hat \pi$, we extract the $.95$ confidence intervals of $\pi$,
the mean and median ranking scores, and the .95 confidence intervals of ranks.
Algorithm \ref{alg:simulate_P} can be used to simulate a $P$. 
Note that each row of the simulated $P$ already sums to $1$.
\begin{figure}[htb]
\centering
\begin{minipage}{.96\linewidth}
\begin{algorithm}[H]\label{alg:simulate_P}
\SetAlgoLined
1. For each pair of $(i,j)$, estimate a 2-parameter beta distribution using $s_{ij}$, $\hat P_{ii}$, $N_{ij}$, and the Wilson intervals\;
2. Use the beta distributions to simulate $S_{ij}$ and $P_{ii}$ for all $i \in \mathbb{N}$ and $j<i$. For $j>i$, let $S_{ij}=1-S_{ji}$\;
3. Apply Algorithm \ref{alg:P_ij} together with simulated $P_{ii}$ and $s_{ij}=S_{ij}$ to calculate $P_{ij}$ whenever $j\not = i$. \
\vspace{.5cm}
\caption{Simulate a $P$.}
\end{algorithm}
\end{minipage}
\end{figure}

\section{Properties of the Sorting}\label{sect:Property}
\noindent
This section discusses a few properties of $\pi$. Some of them relate to the theory of homogeneous Markov chains. Hence, one could use the theory to find more properties.
 Besides, we highlight a few ways to improve a college's ranking score. 

\subsection{Endogenous Weighting}
\noindent
The explicit dimension of the matrix $P$ is $n$.
Moreover, each $P_{ij}$ is a result of choices made by many students who are admitted by both colleges $i$ and $j$;
each student considers an indefinite number of reasons which are not listed in $P_{ij}$.
For any $i \in \mathbb{N}$, we use college $i$ as the benchmark or reference to rank all colleges. 
The ranking scores are a row vector of
$$
\mu_i \ \eqdef \ (P_{i1}, P_{i2}, \cdots, P_{in}).
$$ 
The more college $j$ influences $i$, the higher its ranking score $P_{ij}$.
This way, we define $n$ sets of ranking scores, $\mu_1, \mu_2, \cdots, \mu_n$.
As each college is unique (i.e., not a linear combination of others), each reference college has its own dimensionality. 
Thus, we have $n$ ranking scores on $n$-dimensional axes, in contrast to merely several axes as in the rankings by mainstream media.

In contrast to (\ref{eq:popular_weighting}),
we apply endogenous weighting to the references to combine these ranking scores $\mu_1, \cdots, \mu_n$. 
Essentially, we let colleges judge colleges themselves without any external interventions from business interests, mainstream media, advertising, or college administrators.
In doing so, we believe that more weight should be placed on a good reference (i.e., with large $\pi_i$), and less weight should be placed on a bad one (i.e., with small $\pi_i$).
Thus, we have weighted ranking scores $\sum\limits_{i=1}^n \pi_i \mu_i = \pi P$. 
As the weighted ranking scores also quantify the quality of the schools as references, $\pi P$ should be a multiple of $\pi$.
By Theorem \ref{thm:uniqueness_pi}, there exists a unique equation that links the endogenously weighted scores $\pi P$ to a constant multiple of $\pi$; and that equation is (\ref{eq:authority_distribution}).
When $\pi$ is not assumed to have a unit sum, we can still have the equation $\pi = \pi P$ as long as the sum of $\pi P$ does not deviate from that of $\pi$.

\begin{theorem}[Uniqueness]\label{thm:uniqueness_pi}
If $\pi P = \beta \pi$ for some $\beta>0$, then $\beta=1$.
\end{theorem}

\subsection{Spillover Effects}
\noindent
The distribution $\pi$ takes into consideration the spillover effects of direct influence, making non-competing colleges comparable and smoothing out controversial competitions so that they become consistent.
In a two-step spillover $\pi = \pi P = \pi P^2$, for example, 
$$
\pi_i = \sum\limits_{k=1}^n \sum\limits_{z=1}^n \pi_k P_{kz} P_{zi},
$$
$i$ has a direct influence on $z$ when $P_{zi}>0$ and $z$ has a direct influence on $k$ when $P_{kz}>0$. 
Then, $i$ has an indirect influence on $k$ even if $i$ may have no direct influence on $k$;
$\pi_i$ collects indirect influence from all players in $\mathbb{N}$.
For any integer $m>1$, we may also consider an $m$-step indirect influence using $P^m$. 

There are many similarities between $\pi$ and the invariant measure in the theory of homogeneous Markov chains.
 We borrow the concepts of irreducibility and aperiodicity from the theory. 
In general, $P$ satisfies both irreducibility and aperiodicity if we consider the top 300 U.S. colleges. 
In the preference survey data, for example, there is a chain of direct bilateral influence: Northern Virginia Community College (NVCC) $\leftrightarrow$ George Mason University $\leftrightarrow$ UVA $\leftrightarrow$ Harvard.
Thus, seemingly unrelated Harvard and NVCC establish an indirect influence over each other through spillovers of up to three steps.
Besides, irreducibility and aperiodicity guarantee the convergence in Algorithm \ref{alg:Solve_pi}.
Finally, the solution to
\begin{equation} \label{eq:authority_distribution_m}
\pi = \pi P^m
\end{equation}
is the same as the solution to (\ref{eq:authority_distribution}), as stated in Theorem \ref{thm:invaniance_direct_indirect}.

\begin{theorem}\label{thm:invaniance_direct_indirect}
If $P$ is irreducible and aperiodic, then the solutions to (\ref{eq:authority_distribution}) and (\ref{eq:authority_distribution_m}) are equivalent.
\end{theorem}

 The ranking score $\pi_i$ is college $i$'s long-run influence in the higher education system and across all other colleges.
A straightforward implication of Theorem \ref{thm:invaniance_direct_indirect} is that $\pi$ is invariant to any steps of indirect influence. 
If we treat $P^m$ as a new power transition matrix, then conflicts in $P^m$ are less severe than those in $P$ whenever $m>1$.
As $m\to \infty$, $P^m$ gradually smooths out all conflicts in $P$. 
In the long-run in $P^\infty$, all rows are $\pi$, and there are no more conflicts.
Moreover, the spillover effects take two further actions: amplification of real comparative advantages and off-setting of noised ones.
One consequence is that some elite schools have substantial authority compared to non-elite ones.

\subsection{Strategies to Improve $\pi_i$}
\noindent
For college $i$, a policy implication from the ranking is how to boost its relative strength in the higher education system.
Both the college and its students can improve $\pi_i$.
Collaboration with another college may also augment its ranking score.

We introduce a few more notations for the next two theorems. 
Let $I_n$ be the $n\times n$ identity matrix and let $\pi_{_{-i}}$ be the transpose of $\pi$ with $\pi_i$ removed.
The column vector $\alpha_i$ takes the $i$th row of $P$ and then drops its $i$th element, and the matrix $Z_i$ is the transpose of $P$ with the $i$th row and the $i$th column removed.
Also, the column vector $\gamma_{ij}$ extracts the $j$th row from $P$ with the $j$th element replaced with zero, and the $i$th element dropped.

\subsubsection{Unilateral Strategies}
\noindent
At the institutional level, college $i$ could improve its $\pi_i$ by increasing its enrollment rate $P_{ii}$. 
A university can attract its admitted students by such recruiting strategies as tuition discounts, research opportunities, and scholarships.
It could also market its reputation through college sports and alumni networks.
We should not, however, use the single criterion $P_{ii}$ to rank the colleges
---a college could artificially inflate its enrollment rate by simply admitting non-competitive applicants.
Theorem \ref{thm:P_ii} specifies the exact effect of a small variation of $P_{ii}$ on $\pi$.
Derivative (\ref{eq:pi_i_P_ii}) measures the response multiplier of $\pi_i$ for a given small shock of $P_{ii}$.
Not surprisingly, by (\ref{eq:pi_mi_Pii}), rising $P_{ii}$ has a non-positive effect on $\pi_j$ for all $j\not = i$.

\begin{theorem}\label{thm:P_ii}
\begin{equation} \label{eq:pi_i_P_ii}
\frac{\mathrm{d} \pi_i}{\mathrm{d} P_{ii}}
= 
\frac{\pi_i}{1-P_{ii}} \
\frac{\mathbf{1}_{n-1}' (I_{n-1}-Z_i)^{-1} \alpha_i}{1+\mathbf{1}_{n-1}' (I_{n-1}-Z_i)^{-1} \alpha_i} \ge 0
\end{equation}
and
\begin{equation} \label{eq:pi_mi_Pii}
\frac{\mathrm{d} \pi_{_{-i}}}{\mathrm{d} P_{ii}} 
=
- \  \frac{\pi_i}{1-P_{ii} \
\frac{\left (I_{n-1}-Z_i \right )^{-1} \alpha_i}{1+\mathbf{1}_{n-1}' (I_{n-1}-Z_i)^{-1} \alpha_i}.}
\end{equation}
Therefore, $\frac{\mathrm{d} \pi_j}{\mathrm{d} P_{ii}} \le 0$ for all $j\not = i$.
\end{theorem}

At the student level, students could share their private information about college choices on the preference survey websites.
A higher $\pi_i$ is in line with the interest of any student who is committed to enrolling in college $i$.
The student has private information about his or her personal choices, e.g., enrolling in college $i$ and rejecting college $j$. 
Thus, he or she has a strategy to reveal or not the private information on the survey websites.
If the private information is released, then the $j$th row of $P$ changes slightly as $P_{ji}$ increases, but $P_{jj}$ remains the same.
As each choice is counted in $\pi$, both the school and the student are better off if the private information is revealed, according to (\ref{eq:pi_i_P_ji}) in Theorem \ref{thm:P_ji}.
Thus, there is no incentive for students to hide the private information, i.e., strategyproofness.
Derivative (\ref{eq:pi_i_P_ji}) also implies that college $i$ improves $\pi_i$ more when recruiting students from a more competitive college $j$ (i.e., with a more significant $\pi_j$), other things remaining constant.

\begin{theorem}[Strategyproofness]\label{thm:P_ji} For any $j\not = i$,
\begin{equation}\label{eq:pi_i_P_ji}
\frac{\mathrm{d} \pi_i}{\mathrm{d} P_{ji}}
= 
\frac{\pi_j}{1-P_{ji}-P_{jj}} \frac{\mathbf{1}_{n-1}' (I_{n-1}-Z_i)^{-1} \gamma_{ij}}{1+\mathbf{1}_{n-1}' (I_{n-1}-Z_i)^{-1} \alpha_i} \ge 0
\end{equation}
and 
\begin{equation}\label{eq:dPi_dPji}
\frac{\mathrm{d} \pi_{_{-i}}}{\mathrm{d} P_{ji}}
= 
\frac{\pi_j}{1-P_{ji}-P_{jj}} (I_{n-1}-Z_i)^{-1}\left [ \frac{\mathbf{1}_{n-1}' (I_{n-1}-Z_i)^{-1} \gamma_{ij}}{1+\mathbf{1}_{n-1}' (I_{n-1}-Z_i)^{-1} \alpha_i} \alpha_i - \gamma_{ij} \right ] .
\end{equation}
\end{theorem}

\subsubsection{Bilateral Cooperation}
\noindent
The counterbalance equilibrium has a mixed cooperative and non-cooperative character.
From the cooperative side, player $i$ would assist $j$ in improving $\pi_j$ whenever $P_{ji}>0$
because $\pi_jP_{ji}$ is a component of $\pi_i = \sum\limits_{k=1}^n \pi_k P_{ki}$.
From the non-cooperative side, as $\pi \mathbf{1} = 1$, an increase of $\pi_j$ may mean a decrease of $\pi_i$.
 Thus, the trade-off is how much player $i$ should assist player $j$ without sacrificing himself or herself.

 A relevant policy question is how to identify the cooperators and the competitors for player $i$.
By (\ref{eq:dPi_dPji}), $\frac{\mathrm{d} \pi_j}{\mathrm{d} P_{ji}}$ may also be positive. 
If this happens, then colleges $i$ and $j$ would form a cooperative partnership to improve their relative strength by slightly increasing $P_{ji}$.
Formation of this partnership does not involve a third party, so it is easily enforceable.
Once the partnership forms, (\ref{eq:dPi_dPji}) calculates the effects on third parties.
Besides, a third party, say, college $k$, automatically acts as a battlefield for the competition between colleges $i$ and $j$.
The reason is that both $\frac{\mathrm{d} \pi_i}{\mathrm{d} P_{ki}}$ and $\frac{\mathrm{d} \pi_j}{\mathrm{d} P_{kj}}$ are non-negative according to (\ref{eq:pi_i_P_ji}), but $P_{ki}$ directly conflicts with $P_{kj}$.

\section{Results From a Miniature Ranking} \label{sect:empirical}
\noindent
It is not our intention to generate a new college ranking to compete with the commercial ones. 
Also, data collection and computational costs are high if we sort all of approximately $3,000$ U.S. colleges. 
Any truncation of the complete list, however, distorts the ranking results to some degree.
A moderate-sized ranking, based on the above methodology, should be in real-time and online, changing as often as the data of revealed preferences.
To illustrate the methodology in the last sections, however, we analyze the twelve colleges in Table \ref{tab:SamplePreference}, using the data from the websites mentioned above.
Table \ref{tab:Rank_score} reports the ranks and their .95 confidence bands, ranking scores and their .95 confidence bands, and means and medians of $20,000$ simulated sets of ranking scores.

\begin{table}[h]
\centering
\footnotesize 
\caption{A Mini Sample College Ranking}
\label{tab:Rank_score}
\small
\begin{tabular}{r||c|c||c|c||c|c} \hline \hline 
College&Rank&.95 CI&Score $\hat \pi_i$&.95 CI$^*$&Mean $\hat \pi_i$&Median $\hat \pi_i$\\ \hline
Harvard &1 &[1,1] &.2770&[.2757, .2784]&.2770&.2770 \\
Stanford &2 &[2,2] &.2037&[.2024, .2049]&.2037&.2036 \\
Yale   &4 &[4,4] &.1149&[.1142, .1155]&.1148&.1148 \\
Princeton&5 &[5,5] &.0851&[.0844, .0858]&.0851&.0851 \\
MIT   &3 &[3,3] &.1358&[.1347, .1368]&.1358&.1358 \\
Chicago &7 &[7,7] &.0434&[.0430, .0439]&.0434&.0434 \\
CalTech &6 &[6,6] &.0574&[.0567, .0583]&.0574&.0574 \\
Berkeley &8 &[8,9] &.0239&[.0236, .0243]&.0239&.0239 \\
Michigan &9 &[8,9] &.0235&[.0230, .0239]&.0235&.0235 \\
UCLA   &10&[10,10]&.0184&[.0181, .0188]&.0184&.0184 \\
UVA   &11&[11,11]&.0090&[.0087, .0094]&.0090&.0090 \\
UNC   &12&[12,12]&.0079&[.0075, .0084]&.0079&.0079 \\ \hline \hline
\multicolumn{7}{l}{* The .95 confidence intervals (CI) are non-symmetric about $\hat \pi_i$.} \\
\end{tabular}
\end{table}

 As a summary of Table \ref{tab:Rank_score}, the elite private colleges are far ahead of their elite public peers.
Harvard and Stanford capture nearly half of the total authority, due to  the spillover's amplification effect.
 Because of this effect, for example, Harvard's ranking score is 40\% higher than Stanford's while their odds ratios are much closer.
The amplification comes from Harvard's relative advantages over other colleges, compared to Stanford's.
Berkeley and Michigan have very close ranking scores; their confidence intervals overlap by 23\%.
To separate them effectively, we could add more colleges to the ranking.
We could also perform secondary analyses based on the estimated ranking scores.
Regression of the scores on the tuitions, for example, finds which universities are undervalued and which are overvalued.

The estimated power transition matrix $\hat P$ is
{\fontsize{8}{9}\selectfont
$$
\left [
\begin{array}{cccccccccccc}
\mathrm{HRV}&\mathrm{SFD}&\mathrm{Yale}&\mathrm{PRT}&\mathrm{MIT}&\mathrm{CCG}&\mathrm{CIT}&\mathrm{UCB}&\mathrm{Mich}&\mathrm{UCLA}&\mathrm{UVA}&\mathrm{UNC}\\
.7942&.0464&.0411&.0257&.0242&.0167&.0128&.0100&.0132&.0069&.0062&.0026\\
.0723&.7581&.0402&.0260&.0266&.0158&.0338&.0079&.0090&.0046&.0024&.0032\\
.1070&.0634&.6753&.0299&.0468&.0196&.0249&.0046&.0148&.0067&.0035&.0036\\
.1059&.0640&.0514&.6403&.0426&.0215&.0401&.0086&.0089&.0073&.0054&.0040\\
.0616&.0721&.0281&.0340&.6956&.0177&.0543&.0099&.0099&.0089&.0046&.0033\\
.0765&.0560&.0905&.0528&.0689&.5221&.0434&.0160&.0446&.0160&.0056&.0076\\
.0849&.1093&.0288&.0498&.2050&.0329&.3972&.0345&.0210&.0149&.0114&.0104\\
.0407&.0786&.0522&.0432&.0582&.0377&.0993&.4174&.0229&.1428&.0040&.0030\\
.0823&.0582&.0544&.0456&.0787&.0867&.1033&.0428&.4061&.0232&.0093&.0094\\
.0349&.0577&.0344&.0307&.0795&.0344&.1508&.1564&.0229&.3907&.0042&.0033\\
.0563&.0617&.0443&.1563&.0628&.0714&.0682&.0219&.0232&.0152&.3856&.0330\\
.0482&.0423&.0464&.1147&.0218&.0369&.0571&.0247&.0405&.0206&.0702&.4767\\
\end{array}
\right ].
$$
}

\noindent
Compared with $s_{ij}$ in Table \ref{tab:SamplePreference}, $\hat P$ also takes the sizes $N_{ij}$, $E_i$, and $A_i$ into consideration. 
The matrix illustrates which college has the most direct influence on college $i$.
In the first column, for example, Harvard pulls in significant authority from other educational superpowers such as Yale, Princeton, CalTech, and Michigan, in the decreasing order of $P_{i1}$.
By the tenth row, Berkeley, CalTech, MIT, and Stanford are the top four influencers on UCLA.
The direct influence on UNC is the most evenly distributed, possibly due to its long distance from the other institutions.
Duke University may be added to the list to break the approximate evenness.

Using (\ref{eq:pi_i_P_ii}), we calculate the response of $\pi_i$, in percentage, to the 1\% shock of the enrollment rate $P_{ii}$, i.e., $\frac{P_{ii}}{\pi_i} \frac{\mathrm{d} \pi_i}{\mathrm{d} P_{ii}} = \frac{\mathrm{d} \log \pi_i}{\mathrm{d} \log P_{ii}} $. 
The result is the following vector

{\fontsize{8}{9}\selectfont
$$
\left [
\begin{array}{cccccccccccc}
\mathrm{HRV}&\mathrm{SFD}&\mathrm{Yale}&\mathrm{PRT}&\mathrm{MIT}&\mathrm{CCG}&\mathrm{CIT}&\mathrm{UCB}&\mathrm{Mich}&\mathrm{UCLA}&\mathrm{UVA}&\mathrm{UNC}\\
2.79&2.496&1.841&1.629&1.975&1.045&.621&.699&.668&.629&.622&.904
\end{array}
\right ].
$$
}

\noindent
Among the private elites, CalTech has the lowest percentage increase of $\pi_i$ with respect to a 1\% hike in the enrollment rate.
Among the public elites, UNC has the highest sensitivity of $\pi_i$ to the increase of the enrollment rate.
For Michigan to catch up with Berkeley, for example, its $\pi_i$ needs a $\frac{.0239}{.0235}-1=1.7\%$ increase. 
This can be achieved by a $\frac{1.7}{.668}=2.55\%$ increase in the enrollment rate, other things remaining unchanged.
Being situated at the top of the ranking, Harvard may have no incentive to raise its enrollment rate, though it also leads in the impulse response vector.
Also, the response function has a wide range, from $.621$ to $2.79$.
If an econometric model is used to model the ranking scores by enrollment rates and other covariates, then the unknown coefficient for enrollment rates would presumably be a constant across all the colleges, which is highly artificial, as shown in the above vector.
Lastly, one could also use (\ref{eq:pi_mi_Pii}) to find the response of $\pi_j$, in percentage, to the 1\% shock of $P_{ii}$.

To locate the partners for college $i$, we look for all $j\not= i$ for which $\frac{\mathrm{d} \pi_j}{\mathrm{d} P_{ji}}\ge 0$, according to Theorem \ref{thm:P_ji}.
We list these partners, calculated from (\ref{eq:pi_i_P_ji}) and (\ref{eq:dPi_dPji}), in the following matrix:

{\fontsize{8}{9}\selectfont
$$
\left [
\begin{array}{cccccccccccc}
\mathrm{HRV}&\mathrm{SFD}&\mathrm{Yale}&\mathrm{PRT}&\mathrm{MIT}&\mathrm{CCG}&\mathrm{CIT}&\mathrm{UCB}&\mathrm{Mich}&\mathrm{UCLA}&\mathrm{UVA}&\mathrm{UNC}\\
2.79&&\frac{.269}{.013}&\frac{.211}{7e-3}&&&&&&&&\\
&2.50&&\frac{.157}{3e-4}&\frac{.114}{4e-3}&&\frac{.195}{.011}&\frac{.120}{1e-3}&\frac{.129}{3e-4}&\frac{.086}{5e-4}&\frac{.088}{7e-5}&\\
\frac{.159}{1e-4}&&1.84&\frac{.103}{2e-3}&&\frac{.105}{6e-3}&&\frac{.039}{2e-4}&\frac{.120}{1e-3}&&\frac{.073}{1e-4}&\frac{.099}{2e-4}\\
&&&1.63&\frac{.078}{7e-4}&\frac{.085}{3e-3}&\frac{.097}{.004}&\frac{.055}{6e-4}&\frac{.054}{8e-4}&\frac{.058}{3e-4}&\frac{.083}{3e-3}&\frac{.081}{2e-3}\\
&&&&1.98&\frac{.112}{2e-3}&\frac{.208}{.044}&\frac{.100}{2e-3}&\frac{.095}{2e-3}&\frac{.111}{3e-3}&\frac{.113}{4e-4}&\\
&&&&&1.05&\frac{.052}{.001}&\frac{.052}{1e-3}&\frac{.137}{.010}&\frac{.064}{9e-4}&\frac{.044}{9e-4}&\frac{.080}{7e-4}\\
&&&&\frac{.256}{.02}&\frac{.088}{1e-4}&.621&\frac{.148}{8e-3}&\frac{.086}{.004}&\frac{.079}{5e-3}&\frac{.118}{8e-4}&\frac{.144}{7e-4}\\
&&&&&&&.699&&\frac{.318}{.061}&&\\
&&&&&\frac{.094}{.009}&&\frac{.074}{9e-4}&.688&\frac{.050}{5e-4}&\frac{.039}{2e-4}&\frac{.053}{7e-4}\\
&&&&&&&\frac{.213}{.060}&&.629&&\\
&&&&&&\frac{.017}{4e-4}&&\frac{.015}{7e-5}&&.622&\frac{.072}{7e-3}\\
&&&&&\frac{.013}{5e-6}&\frac{.012}{3e-4}&&\frac{.023}{3e-4}&&\frac{.100}{6e-3}&.904\\
\end{array}
\right ].
$$
}

\noindent
At the $i$th row and the $j$th column, the numerator is the response of $\pi_i$, in percentage, to the 1\% shock of $P_{ji}$;
the denominator is the response of $\pi_j$, in percentage, to the same shock.
Based on this matrix, the partnership at $(j,i)$ does not automatically imply a partnership at $(i,j)$. 
Many partnerships do exist at both $(j,i)$ and $(i,j)$, for example, (Harvard, Yale), (MIT, CalTech), and (Michigan, Chicago).
Three or more institutions could also form a partnership, such as (Michigan, UVA, UNC) and (CalTech, UVA, UNC).
Lastly, the numbers in the matrix show which partnership is the favorite one. 
In the tenth column, for example, UCLA has the largest response function $.061$ with Berkeley. 
Therefore, it would prefer the cooperation with Berkeley to that with Stanford, Princeton, MIT, or Chicago.

The above accounts of cooperation work only for a small change of $P$. 
For a large perturbation, we could conduct other analyses.
For example, in a scenario analysis in which CalTech hypothetically raises its enrollment rate from the current level 39.72\% to the same level 69.56\% as MIT, other things remaining the same,
the ranking scores for this scenario become

{
\fontsize{8}{9}\selectfont
$$
\left [
\begin{array}{cccccccccccc}
\mathrm{HRV}&\mathrm{SFD}&\mathrm{Yale}&\mathrm{PRT}&\mathrm{MIT}&\mathrm{CCG}&\mathrm{CIT}&\mathrm{UCB}&\mathrm{Mich}&\mathrm{UCLA}&\mathrm{UVA}&\mathrm{UNC}\\
.2761&.1986&.1159&.0842&.1268&.0432&.0737&.0235&.0232&.0185&.0088&.0076
\end{array}
\right ].
$$
}

\noindent
Compared with Table \ref{tab:Rank_score}, CalTech increases its ranking score by 28\% in this scenario. 
The increase, however, is still not enough for CalTech to surpass Princeton.

\section{Discussion}\label{sect:discussion}
\noindent
The authority-based ranking is subject to several vulnerabilities. 
The survey data could contain selection bias.
For example, STEM (science, technology, engineering, and mathematics) students may be more likely to create accounts on the websites of survey data than students majoring in the humanities. 
This bias would result in higher ranks for schools with strong STEM programs, according to Theorem \ref{thm:P_ji}.
Another way selection bias may creep in is through herding behaviors of the consumers of higher education, which compromise their uniqueness. 
In particular, many consumers have a mindset shaped by mainstream media.
As a consequence, the odds ratios in Table \ref{tab:SamplePreference} could have already been distorted by other rankings.
Secondly, one arguable assumption we make is that revealed preferences effectively capture the most important college comparisons.
 Of course, any additional data would help.
Last but not least, when $\hat \pi_i$ and $\hat \pi_j$ are too close to show a significant difference, it would be unfair to rank one higher and the other lower.
This is often the case when schools are ranked below the top 100.
Other information may be needed to distinguish them.
Otherwise, we could only list the ranking's $.95$ confidence range for each college, which ranks beyond the top $100$.

We could apply the authority-distribution methodology in many similar situations. 
One example, as mentioned in Hu and Shapley (2003), is to sort academic journals.
The only context change is that $P_{ij}$ is the proportion of journal $j$ in all citations cited by journal $i$.
Journal sizes can be adjusted by their total citations using (\ref{eq:new_wilson}).
Indeed, researchers (e.g., Baltagi, 1998; Kalaitzidakis, Mamuneas, and Stengos, 2003 and 2011) have used citation data to rank journals.
For another example, we could consider the relative competitiveness of each currency. 
In this case, we let $P_{ij}$ be the proportion of the export to country $j$ in the total production in country $i$, all in local currency. 
This example has a vivid power-in-and-out dynamic. 
Its results can help avoid unnecessary trade wars and identify the best trade partners. 
In sports, a team is ranked in the news media by how many times it wins throughout the season. 
This simple measurement does not account for how strong the opponents were in those games. 
A remedy could use the endogenous weighting system (\ref{eq:authority_distribution}) so that winning over a strong opponent counts more than winning over a weak one.
It could also account for by how much that team wins or loses in each game so that each earned or lost point is reflected in the ranking.
After being converted into percentages of the total points, the earned points versus lost points in a game make a bilateral odds ratio, like those in Table \ref{tab:SamplePreference}.

How to extend the authority-distribution framework remains a big challenge. 
First, one could analyze the properties of matrix $P$, such as its eigenvalues and eigenvectors.
If we do not normalize the rows of $P$, we may solve the equation $\pi P = \beta \pi$ for some unknown $\beta>0$. 
This solution $\pi$ is the eigenvector centrality of the network, which could avoid scaling the enrollment sizes in Section \ref{subsect:Estimate_P}.
In the sample ranking in Section \ref{sect:empirical}, many other universities not listed in the ranking also have direct  impacts on UCLA, for example.
Thus, the tenth row of $\hat P$ should have a sum of less than one while the diagonal of $\hat P$ remains unchanged.
The shrinkage of the off-diagonal sums could solve the data truncation problem when we are only interested in the top $300$ colleges, which have a national reputation.
The shrinkage size may negatively relate to the ranking score $\pi_i$.
 For this shrunk matrix, the Perron-Frobenius theorem (cf. Keener, 1993) asserts the existence of a positive eigenvector.
Secondly, any other data would help provide a more detailed profile of U.S. higher education. 
One could use data from the CDS, the NCES, and Avery et al. (2013) to conduct a multivariate logistic regression to estimate $P$.
Thirdly, one could also study a multiple-dimensional $\pi$. 
Another counterbalance equilibrium could address how colleges admit students, using acceptance and rejection by schools. 
This facet highly correlates with the one we study in Sections \ref{sect:methodology} through \ref{sect:Property} and supplements the other side of the story in the many-to-many matching game.
Ignoring the correlation, the solution to this counterbalance equilibrium measures what types of applicants are competitive in applying for colleges.
However, dealing with qualitative and latent variables, such as recommendation letters, is a big hurdle to cross.
 Additionally, from a policy viewpoint, when $n$ is large, we need a fast algorithm to identify the best cooperators and the worst competitors for each player.
Lastly, a consumer could select multiple choices of alternatives, revealing his or her preference over unselected ones.
In approval voting (e.g., Brams and Fishburn, 1978), for example, a voter can select a few from a shortlist of candidates. 
Then the ranking scores measure the confidence in the candidates by the voters.

\section{Conclusion}\label{sect:conclusion}
\noindent
College rankings have emerged because of public demand  and intense competition among institutions of higher learning.
They have become as necessary as high school education counselors and college campus visits. 
There are numerous college rankings in the United States, the most famous of which are generated by mainstream media outlets. 
These publications are supposed to provide useful guidelines for high school seniors and help them make the best college choices. 
However, researches have found that college rankings often mislead students and distort the landscape of higher education. 
We argue that the multiplicity of these rankings is due to the subjective selection of ranking criteria and the subjective weights on the criteria.
The consequence is to reduce higher education institutions to only one stereotype and to undermine selection diversity. 
In contrast, the starting point of our research is to recognize that each college is different, as is each student.
Each adds inclusiveness value to the educational system.
 Beyond that, choosing a college could embrace all aspects of the big data available for that college.
Of course, challenges exist regarding the volume, variety, and veracity of the data, as well as heterogeneity (cf. Fan, Han, and Liu, 2014).

 This paper deals with these challenges. 
For heterogeneity and variety, we let each college decide the selection criteria and their weights; we also let each student decide his or her considerations and their weights. 
These decisions altogether result in millions of matching games, in which we observe only a small fraction of the selection outcomes.
Also, revealed preference from each selection is a ranking between two colleges, derived from an individual's rational consideration of many factors.
 Hence, the preferences together could summarize the most relevant big data about colleges, eliminating voluminous efforts in data collection and storage for a college ranking agency.
The full preferences revealed in the outcomes, however, warrant no consistent utility function nor complete linear ordering of the colleges. 
To resolve this issue, we apply authority distribution to absorb the spillover effects in the direct bilateral influence. 
The result is unique and consistent;
 the solution mitigates the noise in data quality by offsetting the inconsistencies in direct bilateral comparisons.

Our ranking method is likely the most authoritative one compared with those used in popular rankings.
First, the results are comprehensive. 
They are based on foot voting by millions of students 
---each student uses his or her decision function. 
Secondly, the method is scientific and objective. 
We use weighting on colleges, but the weights are endogenously implied from the weighting system. 
They are not subjectively determined by a committee,
and they are the ranking scores.
Finally, it offers individual rationality and strategyproofness for students.
The ranking counts every rational choice made by students;
 there is no white noise, as in a regression model.

\vskip 1cm
\noindent \section*{References}

\hangindent=1.1em
\hangafter=1
\noindent
1. Academic Ranking of World Universities (ARWU),
\url{www.shanghairanking.com}. Accessed 1 January 2019.

\hangindent=1.1em
\hangafter=1
\noindent
2. Avery CN, Glickman ME, Hoxby CM, Metrick A. 
A revealed preference ranking of U.S. colleges and universities. 
\textit{Quart J Econ}. 2013;128:425-467.

\hangindent=1.1em
\hangafter=1   
\noindent
3. Baltagi BH. 
Worldwide institutional rankings in econometrics: 1989-1995.
\textit{Econometric Theo}. 1998;14:1-43.

\hangindent=1.1em
\hangafter=1   
\noindent
4. Baucells M, Shapley LS.
Multiperson utility.
\textit{Games Econ Behav}. 2008; 62:329-347.

\hangindent=1.1em
\hangafter=1   
\noindent
5. Bastedo MN, Bowman NA. 
The U.S. News and World Report college rankings: modeling institutional effects on organizational reputation. 
\textit{Am J Educ}. 2010;116:163-184.

\hangindent=1.1em
\hangafter=1   
\noindent
6. Bastedo MN, Bowman NA.
College rankings as an interorganizational dependency: establishing the foundation for strategic and institutional accounts.
\textit{Research Higher Educ}. 2011;52:3-23.

\hangindent=1.1em
\hangafter=1   
\noindent
7. Brams SJ, Fishburn PC.
Approval voting.
\textit{Am Polit Sci Rev}. 1978;72:831-847.

\hangindent=1.1em
\hangafter=1   
\noindent
8. Bruni F.
How to make sense of college rankings.
\textit{New York Times}. 29 October 2016.

\hangindent=1.1em
\hangafter=1
\noindent
9. Chirinko RS, Schaller H.
A revealed preference approach to understanding corporate governance problems: Evidence from Canada.
\textit{J Financial Econ.} 2004; 74:181-206.

\hangindent=1.1em
\hangafter=1   
\noindent
10. Common Data Set Initiative (CDS), 
\url{https://www.commondataset.org}. Accessed 1 January 2019.

\hangindent=1.6em
\hangafter=1   
\noindent
11. Corrente S, Grecoa S, Lowinski RS.
Robust ranking of universities evaluated by hierarchical and interacting criteria.
In: Huber S., Geiger M., de Almeida A., editors. Multiple Criteria Decision Making and Aiding.
Cham: Springer; 2018. p.145-192.

\hangindent=1.6em
\hangafter=1
\noindent
12. Craig R.
College disrupted: the great unbundling of higher education.
New York: St. Martin's Press; 2015.

\hangindent=1.6em
\hangafter=1  
\noindent
13. Crama I, Leruth L.
Control and voting power in corporate networks: Concepts and computational aspects.
\textit{Euro J Oper Res}. 2007;178:879-893.

\hangindent=1.6em
\hangafter=1  
\noindent 
14. Dohmen T, Falk A.
Performance pay and multidimensional sorting: productivity, preferences, and gender.
\textit{Amer Econ Rev.} 2011; 101:556-590.

\hangindent=1.6em
\hangafter=1  
\noindent
15. Eeckhout J, Pinheiro R, Schmidheiny K. 
Spatial sorting.
\textit{J Polit Econ.} 2014; 122:554-620.

\hangindent=1.6em
\hangafter=1  
\noindent
16. Eeckhout J, Pinheiro R, Schmidheiny K. 
Spatial sorting: why New York, Los Angeles and Detroit attract the greatest minds as well as the unskilled.
CESifo Working Paper no. 3274, 2010.

\hangindent=1.6em
\hangafter=1   
\noindent
17. Ehrenberg RG.
Method or madness? inside the U.S. News \& World Report College rankings.
\textit{J College Admission}. 2005;189:29-35.

\hangindent=1.6em
\hangafter=1   
\noindent
18. Fan J, Han F, Liu H.
Challenges of big data analysis.
\textit{Nat Sci Rev.} 2014; 1:293-314.

\hangindent=1.6em
\hangafter=1   
\noindent
19. Forbes, \url{www.forbes.com}, retrieved January 1, 2019.

\hangindent=1.6em
\hangafter=1   
\noindent
20. Gale D, Shapley LS.
College admissions and the stability of marriage.
\textit{Amer Math Mon}. 1962;69:9-15.

\hangindent=1.6em
\hangafter=1   
\noindent
21. Grabisch M, Rusinowska A.
A model of influence in a social network.
\textit{Theo Decision}. 2010;69:69-96.

\hangindent=1.6em
\hangafter=1   
\noindent
22. Grewal R, Dearden JA, Lilien GL.
The uiversity rankings game: modeling the competition among universities for ranking.
\textit{Am Stat}. 2008;62:232-237.

\hangindent=1.6em
\hangafter=1   
\noindent
23. Grinold RC, Kahn RN.
Breadth, skill, and time.
\textit{J Portfolio Manag}. 2011;38:18-28.

\hangindent=1.6em
\hangafter=1   
\noindent
24. Hu X, Shapley LS.
On authority distributions in organizations: equilibrium.
\textit{Games Econ Behav}. 2003;45:132-152.

\hangindent=1.6em
\hangafter=1   
\noindent
25. Kalaitzidakis P, Mamuneas TP, Stengos T.
Rankings of academic journals and institutions in economics.
\textit{J Euro Econ Assoc}. 2003;1:1346-1366.

\hangindent=1.6em
\hangafter=1   
\noindent
26. Kalaitzidakis P, Mamuneas TP, Stengos T.
An updated ranking of academic journals in economics.
\textit{Can J Econ}. 2011;44:1525-1538.

\hangindent=1.6em
\hangafter=1   
\noindent
 27. Keener JP.
The Perron–Frobenius theorem and the ranking of football teams.
\textit{SIAM Rev}. 1993; 35:80-93.

\hangindent=1.6em
\hangafter=1   
\noindent 
28. Langville AN, Meyer CD.
Google's PageRank and Beyond: The Science of Search Engine Rankings.
Princeton, NJ: Princeton University Press; 2012.

\hangindent=1.6em
\hangafter=1   
\noindent
29. Luca M, Smith J.
Salience in quality disclosure: evidence from the U.S. News College rankings.
Harvard Business School Working Paper 2011;12-014.

\hangindent=1.6em
\hangafter=1   
\noindent
30. Moed HF.
A critical comparative analysis of five world university rankings.
\textit{Scientometrics}. 2017;110:967-990.

\hangindent=1.6em
\hangafter=1   
\noindent
31. National Center for Education Statistics,
\url{https://nces.ed.gov/collegenavigator/}. Accessed 1 January 2019.

\hangindent=1.6em
\hangafter=1   
\noindent
32. Niche,
\url{www.niche.com}. Accessed 1 January 2019.

\hangindent=1.6em
\hangafter=1   
\noindent
33. Parchment,
\url{www.parchment.com}. Accessed 1 January 2019.

\hangindent=1.6em
\hangafter=1   
\noindent
34. Perez-Pena R, Slotnik DE.
Gaming the college rankings. 
\textit{New York Times}. 31 January 2012.

\hangindent=1.6em
\hangafter=1   
\noindent
35. Princeton Review,
\url{www.princetonreview.com}. Accessed 1 January 2019.

\hangindent=1.6em
\hangafter=1   
\noindent
36. Pritchard R.
Revealed Preference Methods for Studying Bicycle Route Choice $-$ A Systematic Review.
\textit{Int J Environ Res Public Health}. 2018; 15:470-507..

\hangindent=1.6em
\hangafter=1   
\noindent
37. Quacquarelli Symonds (QS),
\url{www.topuniversities.com}. Accessed 1 January 2019.

\hangindent=1.6em
\hangafter=1   
\noindent
38. Samuelson PA. 
Consumption theory in terms of revealed preference. 
\textit{Economica} New Series. 1948;15:243-253. 

\hangindent=1.6em
\hangafter=1   
\noindent
39. Tieskens KF, Van Zantena BT, Schulpa CJE, Verburga PH.
Aesthetic appreciation of the cultural landscape through social media: An analysis of revealed preference in the Dutch river landscape.
\textit{Landscape and Urban Planning} 2018; 177:128-137.

\hangindent=1.6em
\hangafter=1   
\noindent
40. Times Higher Education, \url{www.timeshighereducation.com}. Accessed 1 January 2019.

\hangindent=1.6em
\hangafter=1   
\noindent
41. U.S. News \& World Report, \url{www.usnews.com}. Accessed 1 January 2019.

\hangindent=1.6em
\hangafter=1   
\noindent
42. Walls Street Journal, \url{www.wsj.com}. Accessed 1 January 2019.

\hangindent=1.6em
\hangafter=1   
\noindent
43. Washington Monthly, \url{www.washingtonmonthly.com}. Accessed 1 January 2019.

\hangindent=1.6em
\hangafter=1   
\noindent
44. Wilson EB. 
Probable inference, the law of succession, and statistical inference.
\textit{J Am Stat Assoc}. 1927;22:209-212.

\newpage
\noindent \textbf{\large{Appendix}}

\subsection*{A1. Proof of Lemma \ref{lm:tilde_M_ij}}

\noindent As $\tau=1-s_{ij}$ and $\eta$ is the lower confidence bound in (\ref{eq:wilson}), 
$$
\eta 
=
\frac{2s_{ij} M_{ij} + z^2}{2 M_{ij} + 2z^2} - \frac{z}{2 M_{ij} + 2 z^2} \sqrt{4\tau s_{ij} M_{ij} + z^2}.
$$
Therefore
$$
\sqrt{4\tau s_{ij} M_{ij} + z^2}
=
\frac{2s_{ij} M_{ij} + z^2 - 2\eta (M_{ij} + z^2)}{z} 
=
\frac{2(s_{ij}-\eta)}{z} M_{ij} + (1-2 \eta)z.
$$
Next, we square both sides to get
$$
\begin{array}{rcl}
4\tau s_{ij} M_{ij} + z^2
&=&
\frac{4(s_{ij}-\eta)^2}{z^2} M_{ij}^2 + 4(s_{ij}-\eta)(1-2 \eta) M_{ij} + (1-2 \eta)^2 z^2 
\end{array}
$$
or simply,
$$
\left ( \frac{s_{ij}-\eta}{z}\right )^2 M_{ij}^2 + [(s_{ij}-\eta)(1-2 \eta)-\tau s_{ij}] M_{ij} - \eta (1- \eta) z^2  = 0.
$$
The above quadratic equation of $M_{ij}$ has the solution expressed in Lemma \ref{lm:tilde_M_ij}.

\subsection*{A2. Proof of Theorem \ref{thm:expected_pij}}
\noindent The probability density for $\tilde N_{ij} = \tilde n_{ij}$ for all $j\in \mathbb{N}$ is
$
\frac{(A_i)!}{\prod\limits_j (\tilde n_{ij})!} \prod\limits_j P_{ij}^{\tilde n_{ij}}
$
and the marginal probability density for $\tilde N_{ii}=E_i$ is
$
\frac{(A_i)!}{ (E_i)! (A_i - E_i)!} P_{ii}^{E_i} (1-P_{ii})^{A_i-E_i}.
$
Thus, given $\tilde N_{ii}=E_i=\tilde n_{ii}$, the conditional probability density for $\tilde N_{ij} = \tilde n_{ij}$ for all $j\not=i$ is
$$
\frac{\frac{(A_i)!}{\prod\limits_j (\tilde n_{ij})!} \prod\limits_j P_{ij}^{\tilde n_{ij}}}{\frac{(A_i)!}{(E_i)! (A_i - E_i)!} P_{ii}^{E_i} (1-P_{ii})^{A_i-E_i}}
\ = \
\frac{(A_i-E_i)!}{\prod\limits_{j\not = i} (\tilde n_{ij})!} \prod\limits_{j\not = i} \left (\frac{P_{ij}}{1-P_{ii}} \right )^{\tilde n_{ij}}.
$$
Therefore, the conditional $\tilde N_{ij}$ given $\tilde N_{ii}=E_i$ also has a multinomial distribution, and the parameters are
$A_i-E_i$ and $\frac{P_{ij}}{1-P_{ii}}, \forall j\not = i.$

Let $\lambda$ be the probability with which any student ---admitted by college $i$--- would participate in the survey. Then, the joint conditional probability of
$\tilde N_{ij} = \tilde n_{ij}$ and $N_{ij}=n_{ij}$ for all $j\not = i$ is
$$
\begin{array}{ll}
&\frac{(A_i-E_i)!}{\prod\limits_{j\not = i} (\tilde n_{ij})!} \prod\limits_{j\not = i} \left (\frac{P_{ij}}{1-P_{ii}} \right )^{\tilde n_{ij}} 
\prod\limits_{j\not = i} \left [ \left ( \begin{array}{c} \tilde n_{ij} \\ n_{ij} \end{array} \right )\lambda^{n_{ij}} (1-\lambda)^{\tilde n_{ij} - n_{ij}} \right ] \\

=&
\frac{\left (A_i-E_i\right )! \lambda^{\sum\limits_{j\not = i} n_{ij}} (1-\lambda)^{A_i-E_i -\sum\limits_{j\not = i} n_{ij}}}{\left (\sum\limits_{j\not = i} n_{ij} \right )!
	\left (A_i-E_i -\sum\limits_{j\not = i} n_{ij} \right )!}
\times
\frac{\left (\sum\limits_{j\not = i} (\tilde n_{ij} - n_{ij}) \right )! \prod\limits_{j\not = i} \left ( \frac{P_{ij}}{1-P_{ii}} \right )^{\tilde n_{ij} - n_{ij}}}{\prod\limits_{j\not = i} (\tilde n_{ij} - n_{ij})!} \\
&
\times
\frac{\left (\sum\limits_{j\not = i} n_{ij} \right )! \prod\limits_{j\not = i} \left ( \frac{P_{ij}}{1-P_{ii}} \right )^{n_{ij}}}{\prod\limits_{j\not = i} (n_{ij})!}.
\end{array}
$$
In the above three fractions, the first one is the binomial distribution density for $\sum\limits_{j\not = i}N_{ij}$, the total number of admitted students in the survey excluding those in $E_i$. 
The second one is the multinomial distribution density for $\tilde N_{ij}-N_{ij}$ for all $j\not = i$, the admitted students not in the survey also excluding those in $E_i$.
And the third one is the multinomial density for $N_{ij}$, conditional on $\sum\limits_{j\not = i}N_{ij}$.
Finally, we factor out the first two fractions to get the conditional marginal density of $N_{ij} = n_{ij}$ as 
$
\frac{\left (\sum\limits_{j\not = i} n_{ij} \right )!}{\prod\limits_{j\not = i} (n_{ij})!} 
\prod\limits_{j\not = i} \left [\frac{P_{ij}}{1-P_{ii}} \right ]^{n_{ij}}.
$
This shows that $N_{ij}$, $\forall j\not = i$, have a multinomial distribution, given the observation of $\sum\limits_{k \not = i} N_{ik}$. 
Therefore, $\frac{N_{ij}}{\sum\limits_{k \not = i} N_{ik}}$ has a conditional expectation $\frac{P_{ij}}{1-P_{ii}}$.

\subsection*{A3. Proof of Theorem \ref{thm:uniqueness_pi}}
\noindent We multiply $\mathbf{1}_n$ on both right sides of the equation $\pi P = \beta \pi$ to get 
$
\pi P \mathbf{1}_n = \beta \pi \mathbf{1}_n.
$
As $P \mathbf{1}_n = \mathbf{1}_n$ and $\pi \mathbf{1}_n = 1$, we have
$$
\beta = \beta \pi \mathbf{1}_n = \pi P \mathbf{1}_n = \pi \mathbf{1}_n = 1.
$$

\subsection*{A4. Proof of Theorem \ref {thm:invaniance_direct_indirect}}
\noindent When $P$ is irreducible and aperiodic, $P^t$ converges as $t \to \infty$; and
$
\lim \limits_{t\to\infty} P^t =
\mathbf{1}_n (\zeta_1, \cdots, \zeta_n)
$
for some row vector $(\zeta_1, \cdots, \zeta_n)$.
Therefore, by (\ref{eq:authority_distribution}) and $\pi \mathbf{1}_n=1$,
$$
\pi = \pi P 
=\pi P^2
= \cdots
= \lim \limits_{t\to\infty} \pi P^t 
= \pi
\mathbf{1}_n
(\zeta_1, \cdots, \zeta_n)
= (\zeta_1, \cdots, \zeta_n).
$$
And by (\ref{eq:authority_distribution_m}),
$$
\pi = \pi P^m 
=\pi P^{2m}
= \cdots
= \lim \limits_{t\to\infty} \pi P^{tm} 
= \pi
\mathbf{1}_n
(\zeta_1, \cdots, \zeta_n)
= (\zeta_1, \cdots, \zeta_n).
$$

\subsection*{A5. Proof of Theorem \ref{thm:P_ii}}
\noindent Let $\mathbf{0}_n$ be the $n\times 1$ zero vector.
When we make a small perturbation $\Delta P$ to $P$, the new authority distribution $\pi + \Delta \pi$ satisfies the counterbalance equation of
\begin{equation}\label{eq:delta_authority_distribution}\tag{A.1}
\pi + \Delta \pi = (\pi + \Delta \pi) [P + \Delta P]
\end{equation}
subject to $\Delta P \mathbf{1}_n = \mathbf{0}_n$ and $\Delta \pi \mathbf{1}_n = 0$.
After subtracting $\pi = \pi P$ from (\ref{eq:delta_authority_distribution}), we get
$
\Delta \pi [I_n - P -\Delta P] = \pi \Delta P
$
and its first-order approximation
\begin{equation}\label{eq:linear_equations}\tag{A.2}
\Delta \pi [I_n - P] \approx \pi \Delta P.
\end{equation}

Without loss of generality, let us increase $P_{11}$ by $\Delta P_{11}$ and calculate the effect of the change on $\pi$. 
Then the elements of $\Delta P$ are all zeros except the first row.
By (\ref{eq:Pij_Estimation}), the other elements in the row decrease proportionally. 
Thus, the first row of $\Delta P$ is
$$
\left (\Delta P_{11}, \frac{-\Delta P_{11}}{1-P_{11}} P_{12}, \cdots, \frac{-\Delta P_{11}}{1-P_{11}} P_{1n} \right )
=
\Delta P_{11} \left (1, \frac{-P_{12}}{1-P_{11}}, \cdots, \frac{-P_{1n}}{1-P_{11}} \right ).
$$
By (\ref{eq:linear_equations}), the derivative of $\pi$ with respect to $P_{11}$, i.e. $\frac{\mathrm{d} \pi}{\mathrm{d} P_{11}}$, satisfies
\begin{equation}\label{eq:partial_pi}\tag{A.3}
\frac{\mathrm{d} \pi}{\mathrm{d} P_{11}} \left [ I_n -P \right ] = \pi \frac{\mathrm{d} P}{\mathrm{d} P_{11}} 
= \pi_1 \left (1, \frac{-P_{12}}{1-P_{11}}, \cdots, \frac{-P_{1n}}{1-P_{11}} \right ).
\end{equation}

Let the row vector $\beta_i$ take the $i$th column of $P$ and then drop the $i$th element.
We partition the transpose of $P$ as 
$
P' = \left (
\begin{array}{cc}
P_{11}  & \beta_1 \\
\alpha_1 & Z_1 
\end{array}
\right ).
$
To solve $\frac{\mathrm{d} \pi }{\mathrm{d} P_{11}}$ from (\ref{eq:partial_pi}), we write the augmented matrix for the identity $\frac{\mathrm{d} \pi}{\mathrm{d} P_{11}} \mathbf{1}_n = 0$ and the transpose of (\ref{eq:partial_pi}) as
$
\left[ 
\begin{array}{cc | c}
1     & \mathbf{1}_{n-1}' & 0 \\
1-P_{11} & -\beta_1      & \pi_1 \\ 
-\alpha_1  & I_{n-1} -Z_1    & \frac{-\pi_1}{1-P_{11}} \alpha_1 \\ 
\end{array}
\right].
$
If we multiply 
$
\left [
\begin{array}{ccc} 
1 & 0 & \mathbf{0}_n' \\
0 & 1 & \mathbf{1}_{n-1}' \\
\mathbf{0}_n & \mathbf{0}_n & I_{n-1}
\end{array}
\right ]
$
to the left side of the augmented matrix, then the second row becomes a zero vector. 
After dropping the second row, we get the new augmented matrix of
\begin{equation}\label{eq:simplied_augmented}\tag{A.4}
\left[ 
\begin{array}{cc | c}
1        & \mathbf{1}_{n-1}' & 0 \\
-\alpha_1  & I_{n-1} -Z_1    & \frac{-\pi_1}{1-P_{11}} \alpha_1 \\ 
\end{array}
\right].
\end{equation}
We next multiple
$
\left [
\begin{array}{cc}
1 & -\mathbf{1}_{n-1}' (I_{n-1}-Z_1)^{-1} \\
0 & I_{n-1}
\end{array}
\right ]
$
to the left side of (\ref{eq:simplied_augmented}) to get
\begin{equation}\label{eq:trans_augmented}\tag{A.5}
\left[ 
\begin{array}{cc | c}
1+\mathbf{1}_{n-1}' (I_{n-1}-Z_1)^{-1} \alpha_1 & \mathbf{0}_{n-1}' & \frac{\pi_1}{1-P_{11}} \mathbf{1}_{n-1}' (I_{n-1}-Z_1)^{-1} \alpha_1 \\
-\alpha_1  & I_{n-1} -Z_1    & \frac{-\pi_1}{1-P_{11}} \alpha_1 \\ 
\end{array}
\right].
\end{equation}
Note that $(I_{n-1}-Z_1)^{-1} = I_{n-1}+Z_1 + Z_1^2 + Z_1^3 +\cdots$ has all non-negative elements.
Thus, $\mathbf{1}_{n-1}' (I_{n-1}-Z_1)^{-1} \alpha_1 \ge 0$.
By the first row of (\ref{eq:trans_augmented}), 
$$
\frac{\mathrm{d} \pi_1}{\mathrm{d} P_{11}} = \frac{\pi_1}{1-P_{11}} \frac{\mathbf{1}_{n-1}' (I_{n-1}-Z_1)^{-1} \alpha_1}{1+\mathbf{1}_{n-1}' (I_{n-1}-Z_1)^{-1} \alpha_1} \ge 0.
$$
By the second row of (\ref{eq:trans_augmented}), 
$
-\frac{\mathrm{d} \pi_1}{\mathrm{d} P_{11}} \alpha_1 + \left (I_{n-1}-Z_1 \right ) \frac{\mathrm{d} \pi_{_{-1}}}{\mathrm{d} P_{11}} = \frac{-\pi_1}{1-P_{11}} \alpha_1
$
and thus
$$
\begin{array}{rcl}
\frac{\mathrm{d} \pi_{_{-1}}}{\mathrm{d} P_{11}} 
&=& 
\left (I_{n-1}-Z_1 \right )^{-1} \left [ \frac{\mathrm{d} \pi_1}{\mathrm{d} P_{11}} - \frac{\pi_1}{1-P_{11}} \right ] \alpha_1 \\
&=&
- \frac{\pi_1}{(1-P_{11}) \left [1+\mathbf{1}_{n-1}' (I_{n-1}-Z_1)^{-1} \alpha_1 \right ]} \left (I_{n-1}-Z_1 \right )^{-1} \alpha_1.
\end{array}
$$
Therefore, $\frac{\mathrm{d} \pi_{_{-1}}}{\mathrm{d} P_{11}}$ is a non-positive vector.

\subsection*{A6. Proof of Theorem \ref{thm:P_ji}}
\noindent To show a student's strategyproofness, without loss of generality, we assume that the student enrolls in college $1$ and rejects college $2$.
If he or she reveals the private information in the preference survey, we want the effect of the revelation on $\pi$.

Whether revelation or not, the enrollment rates $P_{11}$ and $P_{22}$ do not change because they are official enrollment data
---the enrollment of the student in college $1$ is already counted in $P_{11}$ and never counted in $P_{22}$.
Though $M_{12}$ and $s_{12}$ change with the revelation, their product $N_{12}$ remains unchanged because it is the number of students in the survey who are admitted by college $1$ and enroll in college $2$.
Thus, by (\ref{eq:Pij_Estimation}), the first row of $P$ does not change with the revelation.

However, the estimated $P_{21}$ increases. 
Let $P_{21}$ have a small change $\Delta P_{21}$.
By (\ref{eq:Pij_Estimation}), $\Delta P$ has non-zero values only in the second row which is
{
\small 
$$
\left (
\Delta P_{21}, 0, \frac{-\Delta P_{21}P_{23}}{1-P_{21}-P_{22}}, \cdots, \frac{-\Delta P_{21}P_{2n}}{1-P_{21}-P_{22}}
\right )
=
\Delta P_{21}
\left (
1, 0, \frac{-P_{23}}{1-P_{21}-P_{22}}, \cdots, \frac{-P_{2n}}{1-P_{21}-P_{22}}
\right ).
$$
}
By (\ref{eq:linear_equations}), the derivative of $\pi$ with respect to $P_{21}$, i.e. $\frac{\mathrm{d} \pi}{\mathrm{d} P_{21}}$, satisfies 
\begin{equation}\label{eq:derivative_P21}\tag{A.6}
\frac{\mathrm{d} \pi}{\mathrm{d} P_{21}} \left [ I_n -P \right ] = \pi \frac{\mathrm{d} P}{\mathrm{d} P_{21}} 
= \pi_2 \left (1, 0, \frac{-P_{23}}{1-P_{21}-P_{22}}, \cdots, \frac{-P_{2n}}{1-P_{21}-P_{22}} \right ).
\end{equation}
In solving $\frac{\mathrm{d} \pi}{\mathrm{d} P_{21}}$ from (\ref{eq:derivative_P21}), the augmented matrix for $\frac{\mathrm{d} \pi}{\mathrm{d} P_{21}} \mathbf{1}_n = 0$ and the transpose of (\ref{eq:derivative_P21}) is
$
\left [ 
\begin{array}{cc | c}
1     & \mathbf{1}_{n-1}' & 0 \\
1-P_{11} & -\beta_1      & \pi_2 \\ 
-\alpha_1  & I_{n-1} -Z_1    & \frac{-\pi_2}{1-P_{21}-P_{22}} \gamma_{12} \\ 
\end{array}
\right ].
$
We apply the same operations as in the proof of Theorem \ref{thm:P_ii} to the matrix to get the new augmented matrix
\begin{equation}\label{eq:pi_wrt_P21}\tag{A.7}
\left[ 
\begin{array}{cc | c}
1+\mathbf{1}_{n-1}' (I_{n-1}-Z_1)^{-1} \alpha_1 & \mathbf{0}_{n-1}' & \frac{\pi_2}{1-P_{21}-P_{22}} \mathbf{1}_{n-1}' (I_{n-1}-Z_1)^{-1} \gamma_{12} \\
-\alpha_1  & I_{n-1} -Z_1    & \frac{-\pi_2}{1-P_{21}-P_{22}} \gamma_{12} \\ 
\end{array}
\right].
\end{equation}
Therefore, by the first row of (\ref{eq:pi_wrt_P21}),
$$
\frac{\mathrm{d} \pi_1}{\mathrm{d} P_{21}}
= 
\frac{\pi_2}{1-P_{21}-P_{22}}
\frac{\mathbf{1}_{n-1}' (I_{n-1}-Z_1)^{-1} \gamma_{12}}{1+\mathbf{1}_{n-1}' (I_{n-1}-Z_1)^{-1} \alpha_1} \ge 0.
$$
By the second row of (\ref{eq:pi_wrt_P21}), 
$
-\frac{\mathrm{d} \pi_1}{\mathrm{d} P_{21}} \alpha_1 + \left (I_{n-1}-Z_1 \right ) \frac{\mathrm{d} \pi_{_{-1}}}{\mathrm{d} P_{21}} = \frac{-\pi_2}{1-P_{21}-P_{22}} \gamma_{12}
$
and thus
$$
\begin{array}{rcl}
\frac{\mathrm{d} \pi_{_{-1}}}{\mathrm{d} P_{21}} 
&=& 
\left (I_{n-1}-Z_1 \right )^{-1} \left [ \frac{\mathrm{d} \pi_1}{\mathrm{d} P_{21}}\alpha_1 - \frac{\pi_2}{1-P_{21}-P_{22}} \gamma_{12} \right ] \\
&=&
(I_{n-1}-Z_1)^{-1}\left [ \frac{\pi_2}{1-P_{21}-P_{22}} \frac{\mathbf{1}_{n-1}' (I_{n-1}-Z_1)^{-1} \gamma_{12}}{1+\mathbf{1}_{n-1}' (I_{n-1}-Z_1)^{-1} \alpha_1} \alpha_1 - \frac{\pi_2}{1-P_{21}-P_{22}} \gamma_{12} \right ]\\
&=&
\frac{\pi_2}{1-P_{21}-P_{22}} (I_{n-1}-Z_1)^{-1}\left [ \frac{\mathbf{1}_{n-1}' (I_{n-1}-Z_1)^{-1} \gamma_{12}}{1+\mathbf{1}_{n-1}' (I_{n-1}-Z_1)^{-1} \alpha_1} \alpha_1 - \gamma_{12} \right ].
\end{array}
$$

\end{document}